\documentclass{article}

\usepackage{microtype}
\usepackage{graphicx}
\usepackage{subcaption}
\usepackage{booktabs} % for professional tables
\usepackage{multirow}
\usepackage{amsmath} 
\usepackage{xcolor} 
\usepackage{pifont}
\newcommand{\xmark}{\ding{55}}
\usepackage{hyperref}

\usepackage[accepted]{icml2026}

\usepackage{amsmath}
\usepackage{amssymb}
\usepackage{mathtools}
\usepackage{amsthm}

\usepackage[capitalize,noabbrev]{cleveref}

\theoremstyle{plain}

\theoremstyle{definition}

\theoremstyle{remark}

\usepackage[textsize=tiny]{todonotes}

\icmltitlerunning{Scalable Event Cloud Network for Event-based Classification}

\begin{document}

\twocolumn[
  \icmltitle{Scalable Event Cloud Network for Event-based Classification}

  % It is OKAY to include author information, even for blind submissions: the
  % style file will automatically remove it for you unless you've provided
  % the [accepted] option to the icml2026 package.

  % List of affiliations: The first argument should be a (short) identifier you
  % will use later to specify author affiliations Academic affiliations
  % should list Department, University, City, Region, Country Industry
  % affiliations should list Company, City, Region, Country

  % You can specify symbols, otherwise they are numbered in order. Ideally, you
  % should not use this facility. Affiliations will be numbered in order of
  % appearance and this is the preferred way.
  \icmlsetsymbol{equal}{*}

  \begin{icmlauthorlist}
    \icmlauthor{Hongwei Ren}{equal,MAIA}
    \icmlauthor{Fei Ma}{equal,GM}
    \icmlauthor{Xiaopeng Lin}{MICS}
    \icmlauthor{Yuetong Fang}{MICS}
    \icmlauthor{Hongxiang Huang}{MICS}
    \icmlauthor{Yue Zhou}{MICS}
    \icmlauthor{Yulong Huang}{MICS}
    %\icmlauthor{}{sch}
    \icmlauthor{Haotian Fu}{MICS}
    \icmlauthor{Ziyi Yang}{MICS}
    \icmlauthor{Youxin Jiang}{MAIA}
    \icmlauthor{Xiangqian Wu}{MAIA}
    \icmlauthor{Bojun Cheng}{MICS}
    %\icmlauthor{}{sch}
    %\icmlauthor{}{sch}
  \end{icmlauthorlist}

  \icmlaffiliation{MAIA}{Research Centre for Multimodal Artificial Intelligence and 
Applications, Faculty of Computing, Harbin Institute of Technology}
  \icmlaffiliation{MICS}{MICS Thrust, Hong Kong University of Science and Technology (Guangzhou)}
  \icmlaffiliation{GM}{Guangdong Laboratory of Artificial Intelligence and Digital Economy (SZ)}
  % \icmlaffiliation{comp}{Company Name, Location, Country}
  % \icmlaffiliation{sch}{School of ZZZ, Institute of WWW, Location, Country}

  \icmlcorrespondingauthor{Bojun Cheng}{bocheng@hkust-gz.edu.cn}

  % You may provide any keywords that you find helpful for describing your
  % paper; these are used to populate the "keywords" metadata in the PDF but
  % will not be shown in the document
  \icmlkeywords{Machine Learning, ICML}

  \vskip 0.3in
]

% this must go after the closing bracket ] following \twocolumn[ ...

% This command actually creates the footnote in the first column listing the
% affiliations and the copyright notice. The command takes one argument, which
% is text to display at the start of the footnote. The \icmlEqualContribution
% command is standard text for equal contribution. Remove it (just {}) if you
% do not need this facility.

% Use ONE of the following lines. DO NOT remove the command.
% If you have no special notice, KEEP empty braces:
% \printAffiliationsAndNotice{}  % no special notice (required even if empty)
% Or, if applicable, use the standard equal contribution text:
\printAffiliationsAndNotice{\icmlEqualContribution}

\begin{abstract}
Event cameras are biologically inspired sensors garnering significant attention from both industry and academia.
Mainstream methods favor frame and voxel representations, which reach a satisfactory performance while introducing time-consuming transformations, bulky models, and sacrificing fine-grained temporal information.
Alternatively, Point Cloud representation demonstrates promise in addressing the mentioned weaknesses, but it has limited scalability in abstracting features of higher spatial resolution and longer temporal sequence events. 
In this paper, we propose a \textbf{S}calable \textbf{N}etwork named SECNet to leverage \textbf{E}vent \textbf{C}loud representation.
% including polarity and a greater number of events downsampling from the scaling of space and time.
SECNet integrates polarity at the structural level by innovating the Event-based Group and Sampling module rather than only at the input level. 
To accommodate the surge in the number of events, SECNet embraces feature extraction in the frequency domain via the Fourier transform.
This approach not only substantially extinguishes the explosion of Multiply Accumulate Operations but also effectively abstracts spatio-temporal features.
We conducted extensive experiments on \textbf{ten} event-based datasets, and substantiate the scalability, effectiveness, and efficiency of SECNet. Our code will be available at: \url{https://github.com/rhwxmx/SECNet_ICML}.
\end{abstract}

\section{Introduction}
Event cameras represent a revolutionary advancement in the field of computer vision \cite{lichtsteiner2008128,gallego2020event}.
Unlike traditional frame-based cameras that capture images at fixed intervals, event cameras operate asynchronously, detecting local changes in illumination that exceed a predetermined threshold and generating events independently at the pixel level \cite{posch2010qvga}. 
This fundamental difference allows them to achieve exceptional performance characterized by high dynamic range, low latency, and low power consumption \cite{mueggler2017event}.
Consequently, event cameras demonstrate superior effectiveness in capturing high-speed motion and rapid scene dynamics, generating outputs in microsecond resolution, which far surpasses the frame rate of conventional cameras \cite{delbruck2013robotic}. 

\begin{figure}
\includegraphics[width=1\linewidth]{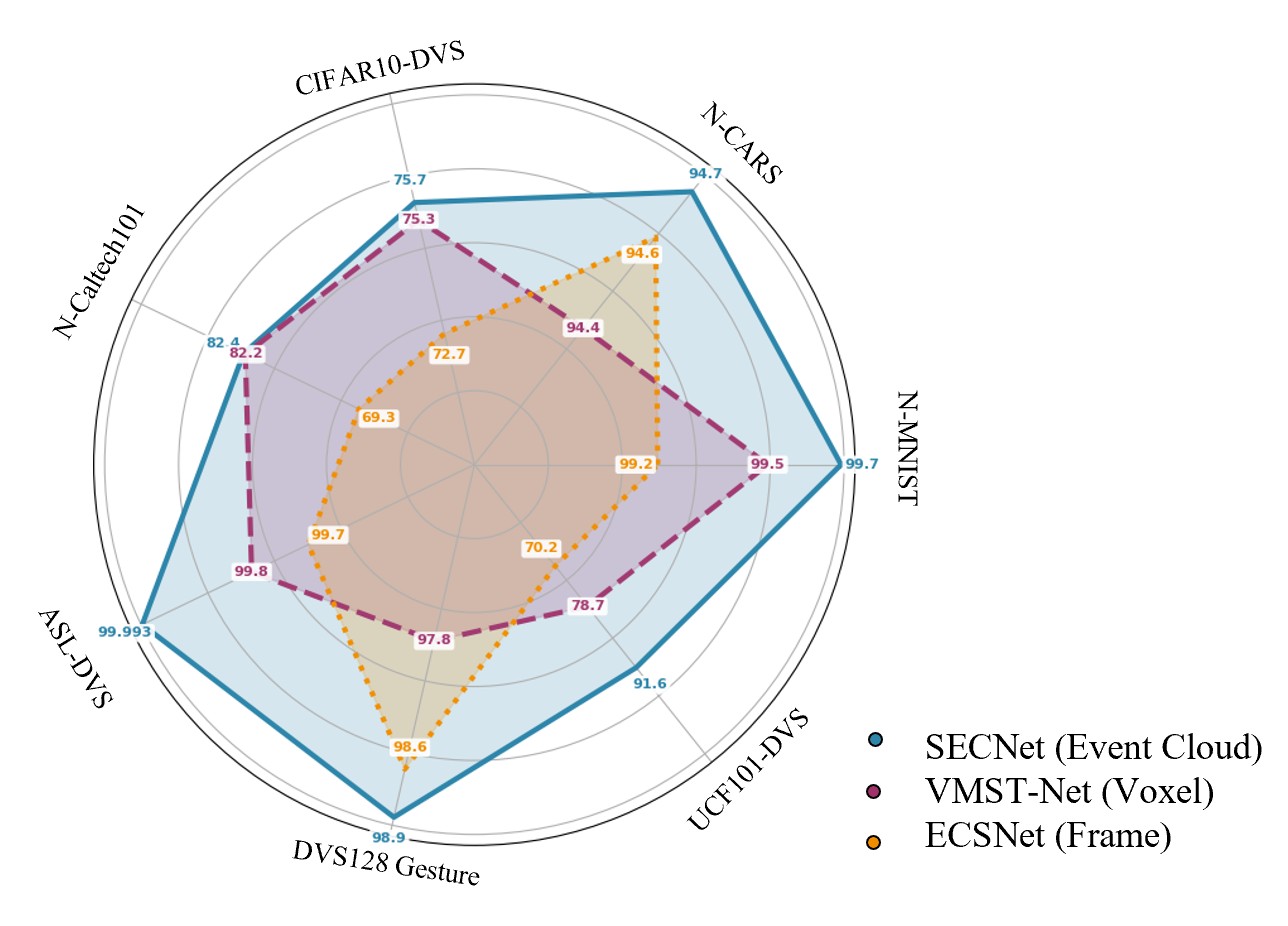}
\caption{ The comparison with the SOTA voxel-based VMST-Net \cite{liu2023voxel} and frame-based methods ECSNet \cite{chen2022ecsnet}  on seven classification datasets.}
\label{fig: gain}
\end{figure}

The raw events generated by event cameras are represented by four dimensions, 2S-1T-1P coordinates, which comprise 2-Spatial coordinate $(x, y)$, 1-Timestamp $t$, and 1-Polarity $p$ of each event occurrence \cite{wang2019space}.
Due to the boom in deep learning algorithms in computer vision, most efforts in event-based vision primarily focus on converting raw events into frame-based or voxel-based representations, employing established backbones to accomplish various tasks, such as VGG \cite{simonyan2014very}, ResNet \cite{he2016deep}, and Transformer \cite{dosovitskiy2020image}. Although these methods generally yield satisfactory performance, they often introduce additional representation transformation times and involve relatively large model architectures, which can substantially diminish their compatibility with hardware and applications \cite{sekikawa2019eventnet}, especially for high-resolution scenes as demonstrated in \cref{table: experiment preprocess}.
On the other hand, the sparseness and fine-grained temporal information contained in raw events are greatly discounted when converted into frames and voxels, which is detrimental for tasks like action recognition that rely on critical details in the temporal dimension \cite{ren2025rethinking}.

Alternative event representation has also been investigated and is gaining popularity. Inspired by PointNet \cite{qi2017pointnet} and PointNet++ \cite{qi2017pointnet++}, several studies employ Point Cloud to represent raw events. Such representation exempts the data format transformation process, preserves the fine-grained temporal information, and is theoretically coupled with lighter network architectures. However, there are \textbf{three major limitations} preventing Point Cloud-based methods from achieving comparable performance to Frame-based and Voxel-based methods. Firstly, Point Cloud representation often ignores the polarity information or merely treats it as part of the input, while treats the temporal coordinate $t$ as a quasi-spatial dimension $z$ \cite{wang2019space,ren2023ttpoint,sekikawa2019eventnet}, failing to differentiate each dimension of events. 
Secondly, the Point Cloud networks show limited scalability due to insufficient extraction of spatial-temporal features in long sequence events from both space and time. Thirdly, the networks' computational load, which is quantified by the Multiply Accumulate Operations (MACs), super-linearly grows with the increase in input points, leading to potential inefficiencies when the input becomes larger \cite{chen2022efficient}; details are provided in \cref{section: super}.

In order to address the constraints mentioned above, we embrace Event Cloud representation \cite{wang2019space}, which is the nearest representation to raw events and requires only downsampling. Compared to the previous Point Cloud method, it gives the polarity and includes multiple times the number of events (1024 to 10240). Then, we innovate a scalable, efficient, and effective framework named SECNet by following the steps. 
Firstly, we redesign the Group and Sampling (G \& S) module to actively incorporate polarity not just at the input level \cite{xie2022vmv,liu2023voxel}, but also into structural decisions such as neighborhood formation and coordinate updates. 
Secondly, we leverage the technique of frequency domain analysis to efficiently extract features in the spatial and temporal domains.  The Spatial Frequency-aware (FA) module significantly reduces the MACs by replacing the convolution with Hadamard’s product by a repetition of $10^4$. 
Meanwhile, the Temporal-FA module excels at capturing long-sequence dependencies of Event Cloud by global frequency filters rather than providing local temporal information. 
Finally, SECNet can proficiently capture the spatio-temporal features with a lightweight architecture scale to higher spatial and longer temporal scenes.
We conducted extensive experiments on ten datasets for three different tasks: object recognition, action recognition, and human pose estimation. The experimental results sufficiently demonstrate the superiority of SECNet, and the gain in seven datasets is shown in \cref{fig: gain}. Our main contribution can be summarized in the following:
\begin{itemize}
    \item SECNet structurally integrates polarity from Event Cloud representation rather than only input level.
    % \item SECNet effectively captures spatial-temporal features from long-term events by integrating time and frequency domains.
    \item SECNet effectively captures spatio-temporal features from long-term events via frequency domain analysis.
    % \item SECNet is capable of handling a greater number of events while being optimized for extremely lightweight.
    \item SECNet is extremely lightweight and scalable to handle high-resolution and long-sequence scenes.
    \item Comparable results highlight that SECNet serves as a powerful backbone for the community.
\end{itemize}

\section{Related Work}
\subsection{Event Representations}
\label{section: representations}
\begin{figure}
\centering
\includegraphics[width=1\linewidth]{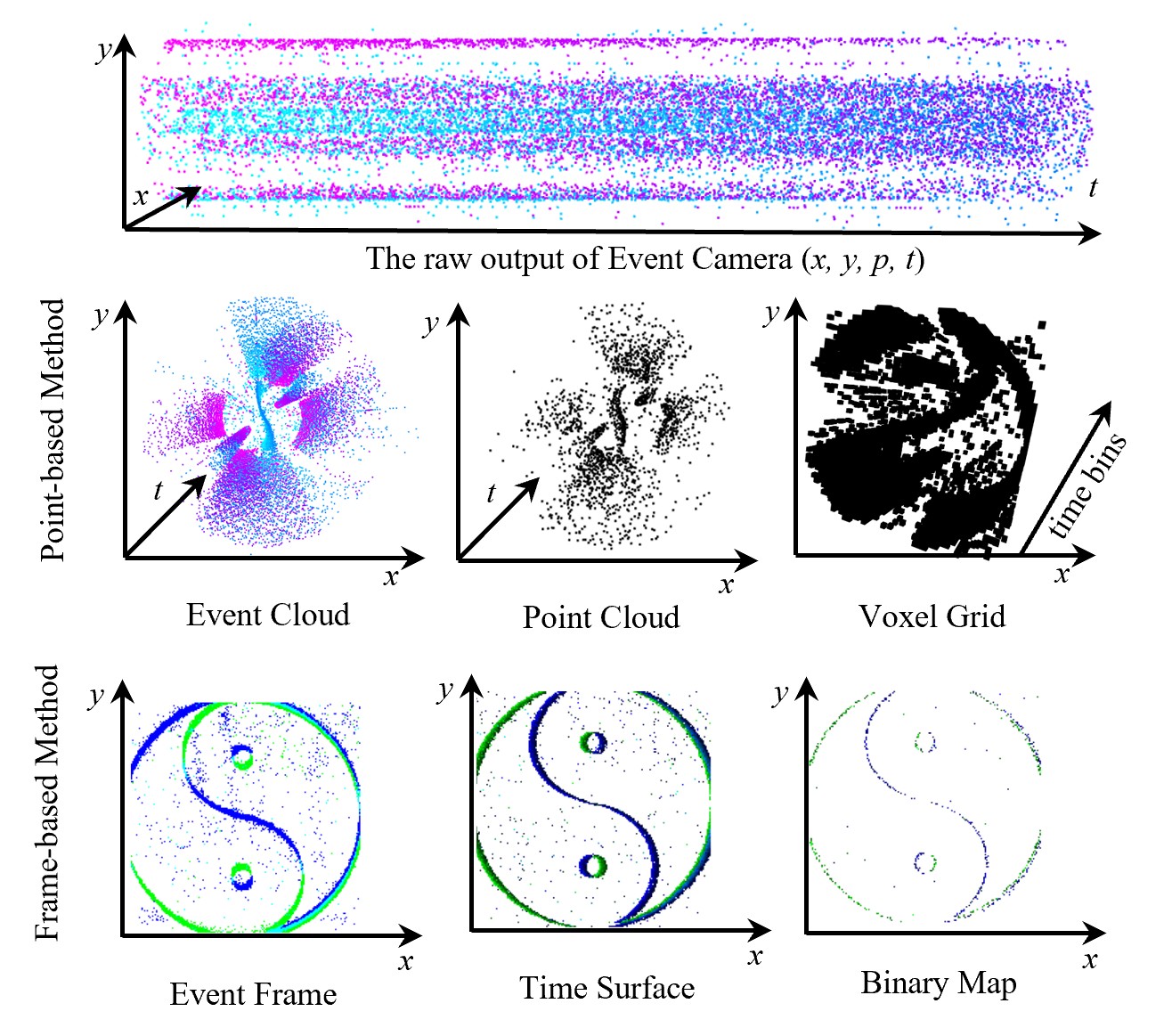}
\caption{Visualization of different representations in yin-yang from N-Caltech101 dataset.}
\label{fig: representations}
\end{figure}
\begin{figure*}[t]
\centerline{\includegraphics[width= 18cm]{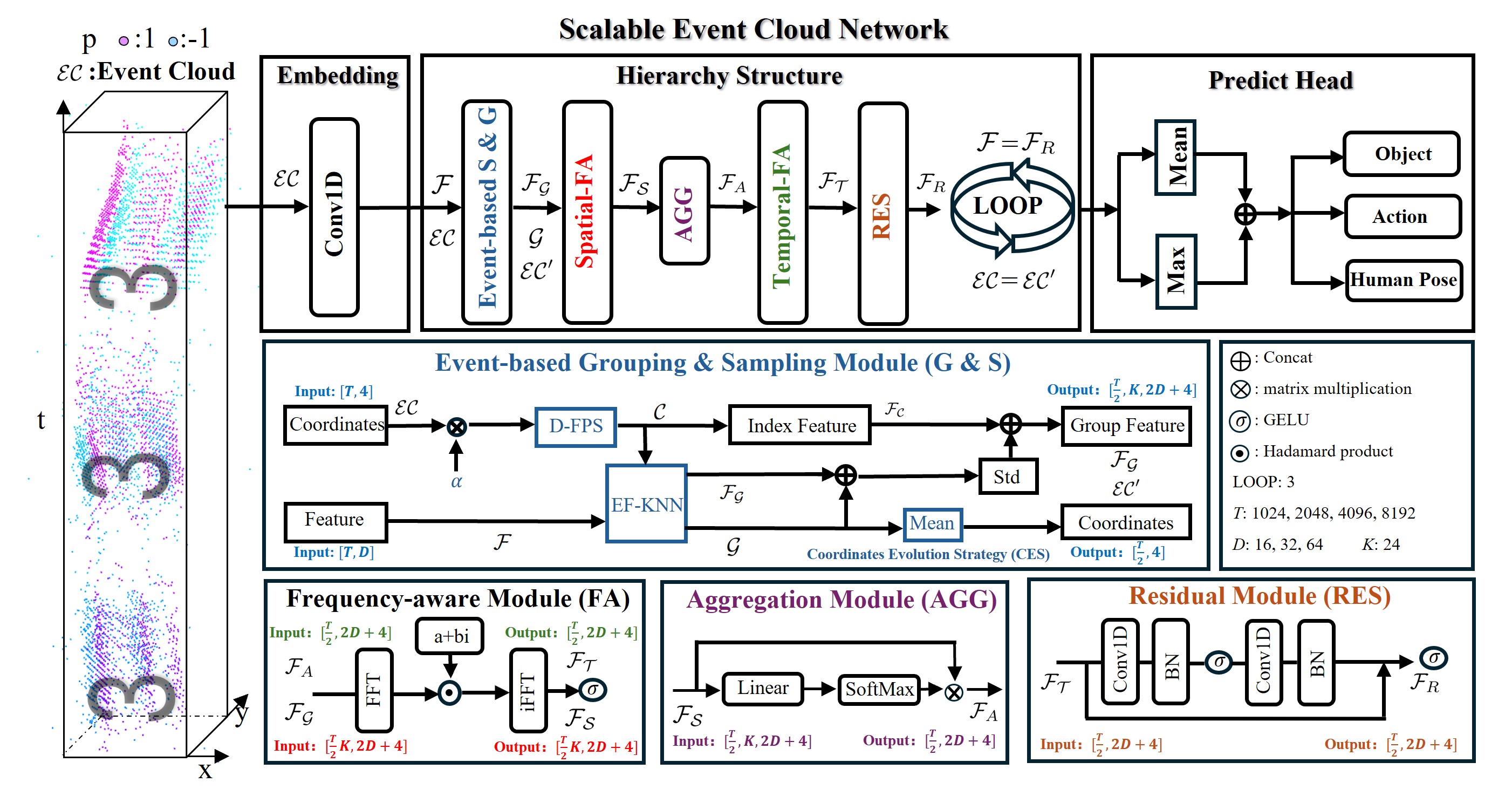}}
\caption{SECNet's architecture. It accomplishes three tasks by processing the Event Cloud into a sequence of distinct modules: Embedding, Hierarchy Structure, and Predict Head. In more detail, the Hierarchy Structure contains five different modules: Event-based G \& S module captures the local neighborhood relationships, Spatial-FA abstracts the explicit spatial and implicit temporal features, AGG aggregates the features in a group, Temporal-FA catches the global explicit temporal features, and RES is responsible for the abstraction and high-level representation of features. The final results are obtained by the max-pooling and average-pooling features.
}
\label{fig: SECNet}
\end{figure*}
Event cameras capture changes in the environment rather than entire frames at regular intervals, allowing them to record rapid movements with high temporal resolution and perform well in challenging lighting conditions. The representation of output from event cameras is a well-developed research problem \cite{gallego2020event}, and we visualize some frame-based and point-based representations in \cref{fig: representations}. The most common one is frame-based representations, which involve converting event streams into 2D frames by either counting the events or summing their polarity over time intervals \cite{gehrig2019end,cannici2020differentiable,lee2016training}. However, this representation often results in a loss of temporal information and sparsity since it compresses dynamic information into static images \cite{wang2019space}. To address this, Time Surface representations are proposed, which encode the timestamp of the most recent event at each pixel location, thereby maintaining some temporal dynamics \citep{sironi2018hats}. Additionally, Bina-Rep converts asynchronous events to a sequence of sparse and expressive event frames \cite{barchid2022bina}, but it is very time-consuming. Another advanced method is voxel-based representation, which segments the event data into multiple time bins, further capturing temporal features across 3D voxel grids or graphs that are beneficial for tasks like 3D reconstruction or depth estimation \cite{xie2022vmv,bi2020graph,deng2022voxel}. Additionally, Point Cloud representation takes a different approach by representing the event data in a 3D spatial format, where each event’s spatial and temporal attributes are preserved \cite{wang2019space,ren2023ttpoint}. A more detailed comparison can be found at \cref{section: Comparison}. However, this representation ignores polarity and generally has a limited point number. Consequently, we embrace Event Cloud representation closest to the raw output, which only needs random downsampling.

\subsection{Frequency-aware Learning}
The Fourier transform is a powerful processing tool in computer vision. With the rapid advancement of convolutional neural networks, the theoretically substitutable frequency-domain Hadamard product offers superior global capture capability and computational efficiency. Initially, FFC \cite{chi2020fast} employed local Fourier units instead of traditional convolutions, performing convolution operations in the frequency domain to process image features at different scales effectively. Subsequently, GFNet \cite{rao2021global} introduced a new technique that enhances feature expression by conducting element-wise multiplication between frequency domain features and learnable global filters. SpectFormer \cite{patro2023spectformer} enhanced the feature representation capabilities of the original ViT \cite{dosovitskiy2020image} architecture by combining spectral analysis with multi-head self-attention mechanisms. Recently, network architectures based on frequency domain modules have been extensively applied to various sub-tasks, such as super-resolution \cite{li2018frequency,chen2024large,xue2020faster}, low-level tasks \cite{mao2023intriguing,mao2024loformer,yu2022frequency,kim2024frequency}, and human pose estimation \cite{zhao2023poseformerv2}. Event Cloud is a collection of long-term signals containing rich spatial and temporal information, and the utilization of frequency-domain feature extraction can better capture the global temporal relationship between events and conduct lightweight networks.

\section{Method}
\subsection{Event Cloud Representation}
Event cameras generate 2-Spatial, 1-Temporal, and 1-Polarity (2S-1T-1P) raw events in response to illumination changes in the environment, which can be defined:
% The raw events $\mathcal{E}$ can be defined as:
\begin{equation}
        \mathcal{E} = \left\{e_i=(x_i,y_i,t_i,p_i) \mid i\in I\right\}, I=[1,2 \ldots, n],
\end{equation}
where $(x, y)$ represents the spatial coordinate where the event emits, $t$ denotes the timestamps, $p$ means the polarity, and $i$ is the index representing $i{\text{-th}}$ element in the event stream.
As described in Section \ref{section: representations}, $\mathcal{E}$ can be transformed into many representations. Still, Event Cloud is the closest representation to raw output and only requires random downsampling. Event Cloud can be generated by:
\begin{equation}
        \mathcal{EC}=\left\{e_j=(x_j,y_j,t_j,p_j) \mid j\in J\right\}, J\subseteq_{\text{ord}} I,
        \label{eq: event cloud}
\end{equation}
where $J$ is a sequential subset of $I$, and the length of $J$ is the number of downsampling points. We define the length of $J$ as $T_0$, so the dimension of $\mathcal{EC}$ is [$T_0$, 4]. Specific settings for each dataset are shown in Table \ref{table: dataset details}. It is crucial to emphasize that Event Cloud does not exhibit the permutation invariance characteristic of Point Cloud. So, the sampled events are arranged in the chronological order of their occurrence, enabling the network to extract explicit events' temporal features subsequently. Finally, Event Cloud is normalized by Min-Max and sent into the network.
\subsection{Overall Architecture}
To fully utilize the spatio-temporal richness and polarity of Event Cloud representation, we design a modular and hierarchical architecture. The overall framework consists of multiple iterative stages.

These operations are applied sequentially to gradually abstract features from sparse, asynchronous events while maintaining computational efficiency. The overall computational flow of the SECNet pipeline is formulated as follows:
\begin{align}
    &\mathcal{G}_i,\mathcal{F}_{\mathcal{G}_i},\mathcal{EC}' = \text{G\&S}(\mathcal{EC},\mathcal{F}_{i-1}),\\
    &\mathcal{F}_{\mathcal{S}_i} = \text{SFA}(\mathcal{F}_{\mathcal{G}_i}), \quad
    \mathcal{F}_{A_i} = \text{AGG}(\mathcal{F}_{\mathcal{S}_i}),\\
    &\mathcal{F}_{\mathcal{T}_i} = \text{TFA}(\mathcal{F}_{A_i}),  \quad
    \mathcal{F}_{R_i} = \text{RES}(\mathcal{F}_{T_i}),\\
    &\mathcal{F}_{i} = \mathcal{F}_{R_i},\mathcal{EC} =  \mathcal{EC}', \text{where }i\in[1,m], \\
    &\mathcal{F}_{\text{end}} = \mathcal{F}_{R_m}, \mathcal{F}_{0} = \text{ Embed}(\mathcal{EC}).
\end{align}
Here, $\mathcal{EC}$ denotes the input Event Cloud, and $\mathcal{F}_i$ represents the feature set at the $i$-th stage, where $\mathcal{F}_0$ is obtained from embedding the initial Event Cloud. $\text{G\&S}(\cdot)$ is the Event-based Grouping and Sampling module that outputs the grouped coordinates $\mathcal{G}_i$, the grouped features $\mathcal{F}_{\mathcal{G}_i}$, and the updated Event Cloud $\mathcal{EC}'$. The Spatial Frequency-aware module is denoted by $\text{SFA}(\cdot)$, yielding spatially enhanced features $\mathcal{F}_{\mathcal{S}_i}$. The attention-based aggregation $\text{AGG}(\cdot)$ then produces $\mathcal{F}_{A_i}$. $\text{TFA}(\cdot)$ represents the Temporal Frequency-aware module, which captures long-range dependencies, producing $\mathcal{F}_{\mathcal{T}_i}$. This is refined through a residual module $\text{RES}(\cdot)$ to obtain the final output $\mathcal{F}_{R_i}$ for stage $i$. At the end of each stage $i$, the feature is updated via $\mathcal{F}_{i} = \mathcal{F}_{R_i}$, and the Event Cloud is replaced by $\mathcal{EC}'$ to reflect the centroid evolution. After $m$ stages, the final feature $\mathcal{F}_{\text{end}}$ is defined as $\mathcal{F}_{R_m}$, which is used for the subsequent classification and regression heads. Next, we elaborate on these innovative modules in the following subsections.
\subsubsection{Event-based Grouping \& Sampling}
Considering the inherent difference in the spatial $(x, y)$, temporal $t$, and polarity $p$ four coordinates within Event Cloud, we redesign the Grouping \& Sampling (G \& S) Module to faithfully extract those by treating them differently and incorporating polarity structurally.
Point-based  G \& S module serves as the fundamental operation that captures local features with the limited receptive field in the Point Cloud space, akin to the convolutional operation in frame-based architecture. Most previous work has ignored the polarity and considered $t$ as $z$ of Point Cloud, where temporal information has been treated as an implicit spatial variable for feature extraction. However, the information represented by 2S-1T-1P is not wholly equivalent to that of Point Cloud. To address this discrepancy, we propose the Event-based G \& S Module, the details of which will be discussed below.

\begin{figure}
\centerline{\includegraphics[width= 7cm]{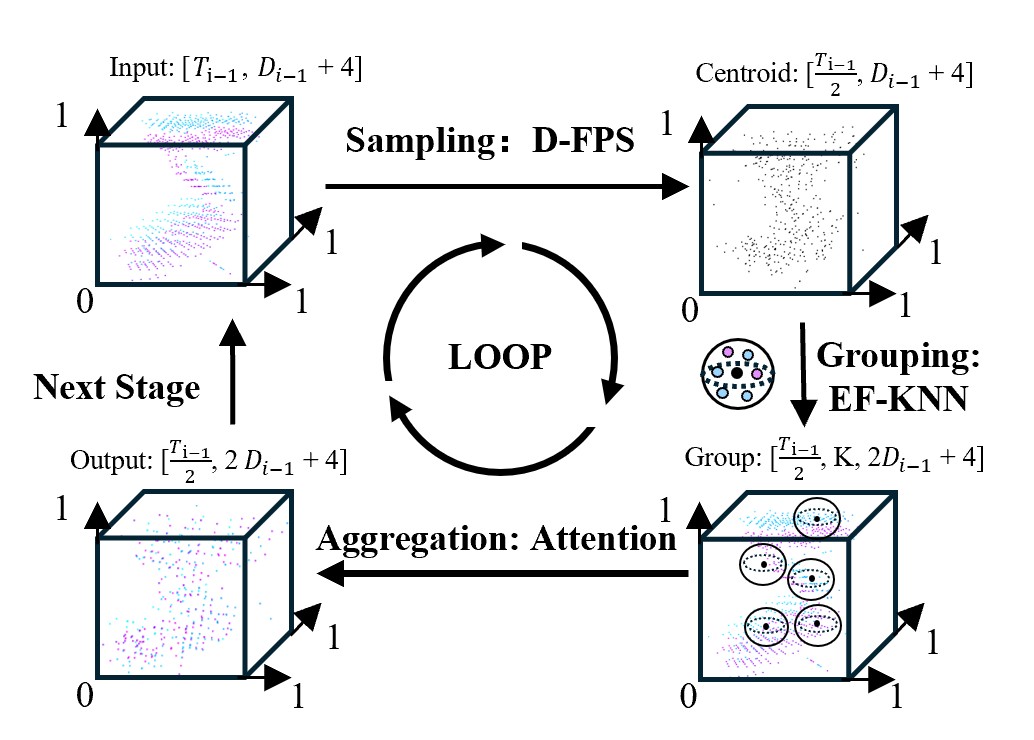}}
\caption{ Visualization of Event Cloud's coordinates and features dimension during hierarchy structure. The closer to the end of the loop, the fewer the number of events and the greater the feature dimension.}
\label{fig: HS}
\end{figure}

Before being fed into the Event-based G \& S module, Event Cloud passes through the embedding layer, as shown in the \cref{fig: SECNet}. The coordinates information represented by $\mathcal{EC}$ with dimension [$T_0$, 4] becomes a high-dimensional feature, defined as $\mathcal{F}_0$, with dimension [$T_0$, $D_0$]. Consequently, the inputs to the Event-based G \& S Module are both  $\mathcal{EC}$ and  $\mathcal{F}_i$. Then, we downsample half events to find the centroid using Farthest Point Sampling (FPS) with the coordinates multiplied by a learnable scaling factor $\alpha$ to focus on different dimensions. We call this process  Differentiation FPS (D-FPS), and this process is formulated by:
\begin{equation}
    \mathcal{C}_i = \text{FPS}(\alpha \cdot \mathcal{EC}), \alpha \in \mathbb{R}^4,\mathcal{C}_i\in \mathbb{R}^{\frac{T_{i-1}}{2}\times 4},
\end{equation}
where $\mathcal{C}_i$ denotes the centroids' coordinates in $i{\text{-th}}$ stage, which are utilized to group events. Then, the centroid $\mathcal{C}_i$ and features $\mathcal{F}_i$  are fed into the  K-Nearest Neighbor (KNN) block \cite{wang2019dynamic}. Different from previous work utilizing coordinates to find the neighbors, we adopt the distance of events' features and call this process Event Features-based KNN (EF-KNN), which is formulated by the following:
\begin{equation}
  \mathcal{G}_i,\mathcal{F}_{\mathcal{G}_i}  = \text{KNN}(\mathcal{F}_{\mathcal{C}_i},\mathcal{F}_i, K),
\end{equation}
where $K$ is the number of events in the group, $\mathcal{F}_{\mathcal{C}_i}$ represents the centroids' feature, $\mathcal{G}_i \in \mathbb{R}^{\frac{T_{i-1}}{2}\times K\times 4}$ denotes the grouped events' coordinate, and $\mathcal{F}_{\mathcal{G}_i} \in \mathbb{R}^{\frac{T_{i-1}}{2}\times K\times D_{i-1}}$ means the grouped events' features which is got by index feature from $\mathcal{F}_i$. The new coordinate is generated by the mean operation within the group, and we call this coordinate update process the Coordinates Evolution Strategy (CES),
\begin{equation}
    \mathcal{EC}' = \text{Mean}(\mathcal{G}_i), \mathcal{EC}' \in \mathbb{R}^{\frac{T_{i-1}}{2}\times 4}.
\end{equation}
The new group feature is concatenated by two elements:
\begin{equation}
    \mathcal{F}_{\mathcal{G}_i} = \mathcal{F}_{\mathcal{G}_i}\oplus\mathcal{G}_i, \mathcal{F}_{\mathcal{G}_i} \in \mathbb{R}^{\frac{T_{i-1}}{2}\times K\times (D_{i-1}+4)},
\end{equation}
where $\oplus$ means the concatenation operation. Finally, the new group feature $\mathcal{F}_{\mathcal{G}_i}$ is obtained under standardized and concatenated with the centroids' features. The standardized process utilizes the group mean feature as the mean value:
\begin{equation}   
\begin{aligned}
\mathcal{M} = \text{Mean}(\mathcal{F}_{\mathcal{G}_i}),&
\mathcal{F}_{\mathcal{G}_i} = \frac{\mathcal{F}_{\mathcal{G}_i}-\mathcal{M}}{\text{Std}(\mathcal{F}_{\mathcal{G}_i},\mathcal{M})},\\
\mathcal{F}_{\mathcal{G}_i} = \mathcal{F}_{\mathcal{G}_i}\oplus\mathcal{F}_{\mathcal{C}_i},&
\mathcal{F}_{\mathcal{G}_i} \in \mathbb{R}^{\frac{T_{i-1}}{2}\times K\times (2D_{i-1}+4)}, 
% \rightarrow \mathbb{R}^{T_i\times K\times D_i}
\end{aligned}
\label{eq: output_g}
\end{equation}
where $\mathcal{M}\in\mathbb{R}^{\frac{T_{i-1}}{2}\times (D_{i-1}+4)}$ denotes the mean values, and Std is the process to compute the standard deviation. With the deepening of the network, coordinates are decreasing and features are increasing, as shown in the \cref{fig: HS}.
\subsubsection{Frequency-aware Module}
We embrace frequency domain analysis to abstract the spatial and temporal features through Spatial Frequency-aware (FA) modules and Temporal-FA modules, respectively.
% Event Cloud contains both spatial and temporal information, and the more events number (T) downsampled, the finer the temporal granularity. 
Event Cloud contains both spatial and temporal information, and the performance is positively correlated to the number of events. Theoretically, more events contain finer spatial and temporal information.
Nevertheless, \textbf{capturing the contextual relationships among a large number of events from space and time poses a challenge for existing event-based network models}, as demonstrated in \cref{table: experiment number}. Simultaneously, the network has to be scalable and efficient to match the low power consumption of the event cameras for a wider range of applications. To address the performance plateau and the energy consumption issue with a large number of events, we adopt the Fourier transform for frequency domain feature extraction, with the process formulated as follows. The definition of spatio-temporal feature can be found \cref{section: st feature}.

Different from images, Event Cloud is a one-dimensional time-series signal [$T_0$, 4], as shown in \cref{eq: event cloud}. One-dimensional discrete Fourier Transform (DFT) is employed to convert the Event Cloud features to the frequency domain by the following formulation:
\begin{equation}
X[k]=\sum_{n=0}^{T-1} x[n] e^{-j \frac{2 \pi}{T} k n}, \quad k=0,1, \ldots, T-1,
\end{equation}
where $j$ is the imaginary unit, $x$ represents the different features ($\mathcal{F}_{\mathcal{G}_i}$ or $\mathcal{F}_{A_i}$) in the hierarchy structure, $X$ denotes the spectrum at different frequencies, and $T$ is the length of temporal signals $x$. Specifically, $\mathcal{F}_{\mathcal{G}_i}$ is reshaped to $[B\times T_i, D_i , K]$, where $B$ is batch size, while $\mathcal{F}_{A_i}$ retains its original shape of $[B, T_i, D_i]$, and the DFT is applied to the $D_i$ spatial dimension and $T_i$ temporal dimensions. Additionally, the inverse DFT can recover the spectrum from temporal signals by the following formulation:
\begin{equation}
x[n]=\frac{1}{T}\sum_{k=0}^{T-1} X[k] e^{j \frac{2 \pi}{T} k n}, \quad n=0,1, \ldots, T-1,
\end{equation}
Mathematically, the $X[T-k] = X^{*}[k]$, where $k\in[0,\frac{T}{2}]$, it means the spectrum $X$ is conjugate symmetric. Therefore, the transformed frequency domain spectrum only needs $\frac{T}{2}+1$ long enough to be recovered to the original signal.

Once the spectrum is obtained through DFT, we can initialize a learnable filter $V$ with dimensionality matching that of the spectrum and perform the Hadamard product between the spectrum and the filter. 
% \begin{equation}
%     \hat{X} = V\odot X
% \end{equation}
In summary, the specific process of the frequency-aware module is as follows:
% \begin{align}
%     X &=  \text{FFT}(x),\\
%     \hat{X} &=   V\odot X,\\
%     x = \text{iFFT}&(\hat{X}),x = \sigma(x),
% \label{eq: fft_process}
% \end{align}
\begin{align}
    X =  \text{FFT}(x),
    \hat{X} =   V\odot X,
    x = \text{iFFT}(\hat{X}), x = \sigma(x),
\label{eq: fft_process}
\end{align}
where $\sigma$ is a nonlinear activation function and $\odot$ means Hadamard product.
As shown in \cref{fig: SECNet}, SECNet adopts two different frequency modules, the Spatial-FA module and the Temporal-FA module. They perform Fourier transforms on Event Cloud features in the spatial $D_i$ and temporal $T_i$ dimensions, respectively, and work on extracting spatial and temporal features. The functions of these two modules will be discussed in detail in the following.
\begin{table*}
\centering
\renewcommand\arraystretch{1}
\scalebox{0.8}{
\begin{tabular}{cccccccccc}
\hline
Dataset        & Source    & Object  & Resolution & Train  & Test  & Point Number & Class & Avg.length (ms)& SW (ms) \\ \hline
NMNIST  \cite{orchard2015converting}       & Synthetic & Static  & 32x32      & 60000  & 10000 & 4096         & 10    & 300        & 300            \\
N-Caltech101 \cite{orchard2015converting}  & Synthetic & Static  & 240x180    & 20896  & 5226  & 8192         & 101   & 300        & 30            \\
CIFAR10-DVS \cite{li2017cifar10}   & Synthetic & Static  & 128x128    & 99965  & 24950 & 10240        & 10    & 1280       & 100            \\
N-Cars   \cite{sironi2018hats}       & Real      & Static  & 128x128    & 15422  & 8607  & 8192         & 2     & 100        & 100            \\
ASL-DVS  \cite{bi2020graph}     & Real      & Static  & 240x180    & 80640  & 20160 & 4096         & 24    & 100        & 100            \\ \hline
DVS128 Gesture \cite{amir2017low} & Real      & Dynamic & 128x128    & 26796  & 6959  & 1024         & 11    & 6520       & 500            \\
Daily DVS \cite{liu2021event}     & Real      & Dynamic & 346x260    & 2924   & 731   & 8192         & 12    & 3000       & 1500           \\
UCF101-DVS \cite{bi2020graph}    & Synthetic & Dynamic & 240x180    & 108065 & 27384 & 8192         & 101   & 6600       & 1000           \\ 
$\text{THU}^{\text{E-ACT}}\text{-50}$ \cite{gao2023action}& Real & Dynamic &1280x800 &62336 &15107 &8192 &50 &- &500\\
\hline
DHP19   \cite{calabrese2019dhp19}       & Real      & Dynamic & 346x240    & 124980       &24588       & 4096         & 13    &    128        &-                \\ \hline
\end{tabular}
}
\caption{Specific details of different datasets. Point Number denotes the fixed downsampled value, and SW represents the sliding window. 
% and SW represents the length of the sampling sliding window. 
% Other included information covers data source, object type, resolution, training/test sample sizes, number of classes, and average duration.
}
\label{table: dataset details}
\end{table*}

\textbf{Spatial-FA Module}. 
% In the realm of signal processing, it is theoretically established that convolution in the temporal domain is equivalent to Hadamard's product in the frequency domain. This equivalence suggests that it is feasible to identify filters in the frequency domain that are mathematically analogous to those derived through backpropagation with the same performance. 
% In the field of image processing, it is theoretically validated that convolution in the spatial domain corresponds to Hadamard's product in the frequency domain after Fourier transformation. This correspondence implies that it is viable to design filters in the frequency domain that are mathematically comparable to those optimized through backpropagation in the spatial domain, achieving equivalent performance.
The input feature $\mathcal{F}_{\mathcal{G}_i}$ to this module is the output of the Event-based G \& S module, with the dimension [$T_i$, K, $D_i$] as shown in \cref{eq: output_g}. 
It represents the topological correlations composed of the spatial proximity, temporal continuity, and polarity relevance of K neighboring points within the group. The low frequencies correspond to globally consistent spatiotemporal patterns within groups, and high frequencies capture local details within neighborhood features.
Previous approaches all employed MLP-based blocks to abstract features across the feature dimensions $D_i$, and this also means doing $T_i\times K$ repeating calculations. The computational complexity of a single MLP is $\mathcal{O}(D_i^2)$. We utilize frequency domain multiplication instead of MLP, and the complexity becomes $\mathcal{O}(D_ilog_{2}(D_i))$. Taking the dimension 540 of the last stage in SECNet, the individual computational complexity is reduced by a factor of 60. This reduced complexity is finally multiplied by the number of repeats $T_i\times K \approx 12\times10^3$. The computational complexity reduction by using the Spatial-FA module is up to 20 times in the actually constructed network. More importantly, the network's performance on the various datasets does not give too much of a discount.

\textbf{Temporal-FA Module}. SECNet adopts a much larger number of input events $T_0$, which is several times as many as in previous Point Cloud-based works. Long-term events have important global and local contextual relationships. Commonly used models like LSTM and attention are not adept at capturing these relationships and would introduce excessive computational complexity, as demonstrated in \cref{table: experiment number}. The input of the Temporal-FA Module is the output of the aggregation module. The aggregation (AGG) module employs a soft attention mechanism on the neighbor dimension to aggregate the feature within a group on the $t$ dimension.
\begin{align}
    A_i = \text{Soft}&\text{Max}(\text{MLP}(\mathcal{F}_{S_i})), A_i\in\mathbb{R}^{T_i \times K},\\
    \mathcal{F}_{A_i} = &A_i \cdot \mathcal{F}_{S_i},\mathcal{F}_{A_i}\in \mathbb{R}^{T_i\times D_i}, 
    % \mathcal{F}_{A_{td}} = \sum_{k=1}^{K} A_{tk} \mathcal{F}_{\mathcal{S}_{tkd}}
\end{align}
where $A_i$ is the attention value of group members, and $\mathcal{F}_{S_i}$ is the output of Spatial-FA module. Then, SECNet performs a Fourier transform in the time dimension $T_i$ of the event features $\mathcal{F}_{A_i}$. And the temporal features of different frequencies are captured through \cref{eq: fft_process}. The filters $V$ are global since they can cover all the frequencies and capture both long-term and short-term interactions. 

In summary, SFA is applied to the grouped feature after the Event-based G \& S, where each group represents a local neighborhood formed by spatial proximity, temporal continuity, and polarity relevance. It can efficiently model intra-group local structure, serving as a lightweight replacement for repeated MLP-based local feature abstraction. In contrast, TFA is introduced after the aggregation stage, when the neighborhood dimension has already been collapsed, and the events remain chronologically ordered. Its role is  to model inter-event temporal dependencies over long event streams, rather than local neighborhood interactions.
\subsubsection{Residual Module}
To abstract the higher-level features within the network from Event Cloud, we employ a basic residual module at the end to further enhance the complexity and hierarchy of features, as shown in \cref{fig: SECNet}. The pseudo code can be found in \cref{section: hardware performance} and specific dimension variance shown in \cref{section: specific dimension}.
% The input and output dimensions of the block are consistent, as shown in \cref{fig: SECNet}.
\begin{table*}[t]
\centering
\renewcommand\arraystretch{1}
\scalebox{0.8}{
\begin{tabular}{ccccccccc}
\hline
Method     & Type  & Publish  & N-MNIST        & N-CARS         & CIFAR10-DVS    & N-Caltech101   & ALS-DVS         & Average\\ \hline
\multicolumn{8}{c}{Pretrained on ImageNet} & \\ \hline
EST  \cite{gehrig2019end}      & Frame & ICCV'19  & 0.991          & 0.925          & 0.749          & 0.837          & 0.991           & 0.899\\
M-LSTM \cite{cannici2020differentiable}    & Frame & ECCV'20  & 0.989          & 0.957          & 0.73           & 0.857          & 0.992           & 0.905\\
MVF-Net \cite{deng2021mvf}   & Frame & TCSVT'21 & 0.993          & 0.968          & 0.762          & 0.871          & 0.996           & 0.918\\ \hline
\multicolumn{8}{c}{Without Pretrainning} & \\ \hline
EST    \cite{gehrig2019end}    & Frame & ICCV'19  & 0.99           & 0.919          & 0.634          & 0.753          & 0.979           & 0.855\\
M-LSTM  \cite{cannici2020differentiable}   & Frame & ECCV'20  & 0.986          & 0.927          & 0.631          & 0.738          & 0.98            & 0.852\\
MVF-Net  \cite{deng2021mvf}   & Frame & TCSVT'21 & 0.981          & 0.927          & 0.599          & 0.687          & 0.971           & 0.833\\
AsyNet  \cite{messikommer2020event}   & Frame & ECCV'20  & -              & 0.944          & 0.663          & 0.745          & -               & -\\
Gabor-SNN \cite{lee2016training} & Frame & CVPR'18  & 0.837          & 0.789          & 0.245          & 0.196          & -               & -\\
HATS  \cite{sironi2018hats}     & Frame & CVPR'18  & 0.991          & 0.902          & 0.524          & 0.642          & -               & -\\
ECSNet  \cite{chen2022ecsnet}   & Frame & TCSVT'22 & 0.992          & 0.946          & 0.727          & 0.693          & 0.997           & 0.871\\
RG-CNNs  \cite{bi2020graph}  & Voxel Graph & TIP'20   & 0.99           & 0.914          & 0.54           & 0.657          & 0.901           & 0.800\\
EV-VGCNN \cite{deng2022voxel}  & Voxel Graph& CVPR'22  & 0.994          & 0.953 & 0.651          & 0.748          & 0.983           & 0.866\\
EVSTr  \cite{xie2024event}    & Voxel Grid& TCSVT'24 & -              & 0.941          & 0.731          & 0.797          & 0.997           & -\\
VMV-GCN \cite{xie2022vmv}   & Voxel Graph& RAL'22   & 0.995          & 0.932          & 0.69           & 0.778          & 0.989           & 0.877\\
EDGCN \cite{deng2024dynamic} &Voxel Graph&AAAI'24 &- &\textbf{0.958} &0.716 &0.801 &-  & -\\
VMST-Net \cite{liu2023voxel} & Voxel Grid& TCSVT'24 & 0.995          & 0.944          & 0.753& 0.822 & 0.998           & 0.902\\
EventNet \cite{sekikawa2019eventnet}  & Point Cloud & CVPR'19  & 0.752          & 0.75           & 0.171          & 0.425          & 0.833           & 0.586\\
PointNet++ \cite{qi2017pointnet++} & Point Cloud & NIPS'17  & 0.841          & 0.809          & 0.465          & 0.503          & 0.947           & 0.713
\\ \hline
SECNet     & Event Cloud & Our      & \textbf{0.997} & 0.947 & \textbf{0.757} & \textbf{0.824} & \textbf{0.9993}  & \textbf{0.905}\\ \hline
\end{tabular}
}
\caption{Experiment on event-based object classification datasets.}
\label{table: experiment object}
\end{table*}
\begin{table*}
\centering
\renewcommand\arraystretch{1}
\scalebox{0.8}{
\begin{tabular}{cccccc}
\hline
Method      & Type & DVSGesture     & DailyDVS       & UCF101-DVS   &Average  \\ \hline
TANet \cite{liu2021tam}      
& F    & 0.974          & 0.963          & 0.669    &0.869      \\
I3D    \cite{carreira2017quo}     
& F    & 0.951          & 0.909          & 0.603   &0.822       \\
ECSNet  \cite{chen2022ecsnet}    
& F    & 0.986          & -              & 0.702     &-     \\
TimeSformer \cite{bertasius2021space}
& F    & 0.917          & 0.906          & 0.541   &0.788       \\
RG-CNN   \cite{bi2020graph}   
& V    & 0.968          & -              & 0.678    &-      \\
VMV-GCN  \cite{xie2022vmv}   
& V    & 0.975          & 0.941          & -        &-      \\
EVSTr \cite{xie2024event} 
&V &0.986 &0.996 &0.735 &0.906\\
VMST-Net \cite{liu2023voxel}   
& V    & 0.978          & -              & 0.787    &-      \\
TTPOINT \cite{ren2023ttpoint}    
& PC    & 0.988          & 0.991          & 0.725   & 0.901      \\
SpikePoint \cite{renspikepoint}  & PC    & 0.988          & 0.979          & 0.685  &  0.884      \\ 
EventMamba \cite{ren2025rethinking} & PC    &\textbf{0.992} &0.991 &0.903 & 0.962 \\\hline
SECNet      & EC    & 0.989 & \textbf{0.9965} & \textbf{0.916} & \textbf{0.967}\\ \hline
\end{tabular}
}
\caption{Event-based action recognition dataset Results.}
\label{table: experiment action}
\end{table*}
\section{Experiment}
\subsection{Implementation Details}
The hardware platform configuration comprises the following: CPU: AMD 7950x, GPU: RTX 4090. Our training model employs the subsequent set of fixed hyperparameters. Optimizer (AdamW), Initial Learning Rate (0.001), Scheduler (Cosine), and Maximum Epochs (150). The Batch Size is dynamically set to take full use of the GPU's memory.

The \textbf{Accuracy} metric used in the object classification and action recognition tasks represents the ratio of correctly predicted samples to the total samples. The \textbf{Mean Per Joint Position Error} (MPJPE) metric is used in the human pose estimation tasks to evaluate the average Euclidean distance between the ground truth and prediction, and the metric space for 2D error is pixels and millimeters for 3D error. The number of skeleton key points set for the experiment is 13. More details can be found in \cref{section: implement details,section: dataset,appendix: comparison method}.
\begin{table}
\centering
\renewcommand\arraystretch{1}
\scalebox{0.9}{
\begin{tabular}{cccc}
\hline
\small Method                 & \small N-Caltech101    & \small Daily DVS    & \small DHP19         \\ \hline
Event Frame           & 4.2ms          & 2.2ms        & 0.5ms          \\
Binary Map         & 772ms          & 286ms        &  717ms              \\
Time Surface           & 4.3ms          & 3.1ms        & 3.3ms          \\
Voxel Grid           & 14.6ms         & 7.9ms        & 4.3ms          \\
Rasterized Point Cloud  & 5.8ms          & 4ms          & 2.9ms          \\ \hline
Event Cloud          & \textbf{1.5ms} & \textbf{1ms} & \textbf{0.3ms} \\ \hline
\end{tabular}
}
\caption{ Preprocessing time of different representations.}
\label{table: experiment preprocess}
\end{table}

\subsection{Preprocessing time}
We utilize tonic \cite{lenz_gregor_2021_5079802} as a code framework, which is a powerful tool for event datasets and transforms. As shown in \cref{table: experiment preprocess}, the time of preprocessing from raw event output to six representations is required on three datasets. It's intuitive to see that Event Cloud takes the least amount of time, and this time barely affects the overall throughput of the model. The Voxel representations, also equipped with lightweight models, spend a lot of time in data format transformation, nearly 10 times more than Event Cloud. This time far exceeds that of network inference, which hinders the practical deployment of the algorithm.

% We validate SECNet on three tasks ten datasets, e.g., \textbf{Object Recognition}: NMNIST  \cite{orchard2015converting}, N-Caltech101 \cite{orchard2015converting}, CIFAR10-DVS \cite{li2017cifar10}, N-CARS   \cite{sironi2018hats}, ASL-DVS  \cite{bi2020graph}, \textbf{Action Recognition}: DVS128 Gesture \cite{amir2017low}, Daily DVS \cite{liu2021event}, UCF101-DVS \cite{bi2020graph}, $\text{THU}^{\text{E-ACT}}\text{-50}$ \cite{gao2023action}, \textbf{Human Pose Estimation}: DHP19 \cite{calabrese2019dhp19}. For N-Caltech101, CIFAR10-DVS, ASL-DVS, Daily DVS, and UCF101-DVS do not have official divisions, we follow the setting in \cite{bi2020graph,chen2022ecsnet}, randomly selected 20\% of the samples as the testset, and all other datasets use official divisions. The details of these datasets are shown in \cref{table: dataset details}. The implementation details and dataset split method can be found in the Appendix.
\subsection{Object Classification Results}
Object classification involves making irregular movements with an event camera around the recognized object, with the classification information primarily encoded in $(x, y)$ coordinates of the event. Theoretically, stacking events into frames would make it easier to reach the performance plateau. But without the pre-training process of ImageNet, the frame-based method performs inferior to the point-based method, which contains the Voxel, Point Cloud, and Event Cloud as shown in \cref{table: experiment object}. This is due to the fact that mapping sparse data onto an intensity 2D plane may have aberrations such as ghosting and blurring, which reduce the quality of the dataset. Prior to our work, Voxel representations dominated such tasks. 
% Point Cloud does not represent objects well due to the limited number of events it supports (1024), resulting in consistently poor performance, which is demonstrated by EventNet and PointNet++. 
Point Cloud has limited scalability to represent objects, resulting in consistently poor performance, which EventNet and PointNet++ demonstrate. 
SECNet adopted Event Cloud representation to improve performance by 27\% compared to the state-of-the-art (SOTA) Point Cloud method. Even compared with the Voxel-based SOTA methods, SECNet shows comparable results on all object classification datasets. 

\subsection{Action Recognition Results}
Action recognition is a task used to analyze human movements or gestures. It is widely applied in various fields, including sports, security, and entertainment. This task differs from object classification in that the main information is embedded in the $t$ coordinates, which represent the type of motion of the observed person or hand. However, voxel-based and frame-based representations tend to diminish the resolution of 
$t$, reducing it from a fine-grained range (1 $\mu s$) to a coarse-grained range (50 $ms$). So, the finer-grained temporal resolution of the representations, the better the performance results, as shown in \cref{table: experiment action}. SECNet utilizes the finest-grained representation of Event Cloud to fully extract its spatial-temporal features, achieving SOTA results on average across three action recognition datasets. Especially on the UCF101-DVS dataset, SECNet far outperforms the previous SOTA methods by 16\%, named VMST-Net. \textbf{This implies that critical information you omit in the representations, and you cannot recover or make it up through the network architecture}. Additionally, we show the hardware performance in UCF101-DVS in \cref{table: ucf hardware}.
\begin{table}
\centering
\scalebox{0.8}{
\begin{tabular}{ccccc}
\hline
Method   & HMAX SNN   & Motion SNN  & EV-ACT & SECNet  \\ \hline
Accuracy & 65.1     & 85.3       & 92.7   & \textbf{97.25} \\ \hline
\end{tabular}}
\caption{
% Experiment on High-resolution $\text{THU}^{\text{E-ACT}}$-50 dataset.
Experiment on  $\text{THU}^{\text{E-ACT}}$-50 dataset.
}
\label{table: experiment thu}
\end{table}
\begin{table}
\centering
\renewcommand\arraystretch{1}
\scalebox{0.8}{
\begin{tabular}{cccccc}
\hline
Name       & Type & Param. & GMACs & FPS  & Acc.  \\ \hline
C3D      & F    & 78.41  & 19.85 & 18   & 0.472 \\
ResNext-50& F    & 26.05  & 3.23  & 35   & 0.602 \\
3D-ResNet  & F    & 63.5   & 14.99 & -    & 0.579 \\
I3D        & F    & 12.37  & 15.06 & 34   & 0.635 \\
EventLSTM  & F    & 21.4   & -     & -    & 0.776 \\
ECSNet     & F    & -      & 6.12  & 0.36 & 0.702 \\
RG-CNN     & V    & 6.95   & 12.46 & -    & 0.632 \\
EVSTr      & V    & 2.88   & 1.38  & 41   & 0.735 \\
VMST-Net   & V    & 3.61   & 0.31  & 23   & 0.787 \\
TTPOINT & P    & 0.357  & 0.294 & 88   & 0.725 \\
EventMamba & P    & 3.28   & 1.861 & 69   & 0.903 \\ \hline
SECNet     & EC   & 1.126  & 0.48  & 120  & 0.916 \\ \hline
\end{tabular}
}
\caption{Hardware benchmark on UCF101-DVS dataset. We adopt the dataset filename-based partitioning method \cite{ren2025rethinking}.}
\label{table: ucf hardware}
\end{table}

\textbf{Scalability to High-resolution dataset}. SECNet outperforms all other methods by 5\%, as presented in \cref{table: experiment thu}. Moreover, compared to EV-ACT \cite{gao2023action}, SECNet consumes one order fewer hardware resources \underline{(1.27 M/ 0.971 MACs vs 21.3 M/ 14.5 MACs)}. It is also evident that the required computation resource is independent of the resolution, making it particularly suitable for processing high-resolution event camera data. 
\begin{table}
\centering
\renewcommand\arraystretch{1}
\scalebox{0.8}{
\begin{tabular}{cccccc}
\hline
\small Method          & \small Type             & \small MPJPE$_{2D}$ & \small MPJPE$_{3D}$  & \small Param. & \small GMACs \\ \hline
Pose-Res50     
& F            & 5.28  & 59.83  & 34     & 12.91 \\
LeVit-128S     
& F            & 7.68  & 87.79  & 7.87   & 0.2   \\
DHP19          
& F            & 7.67  & 87.9   & 0.22   & 3.51  \\ \hline
VMV-PointTrans 
& V           & 9.13  & 103.23 & 3.67   & 5.98  \\
VMST-Net        
& V            & 6.45  & 73.07  & 3.59   & 0.38  \\ \hline
PointNet        
& RPC & 7.29  & 82.46  & 4.46   & 1.19  \\
DGCNN        
& RPC & 6.83  & 77.32  & 4.51   & 4.91  \\
Point Tansformer  & RPC & 6.46  & 73.07  & 3.65   & 5.03  \\
SECNet          & EC      & \textbf{6.11}  & \textbf{69.89}  & \textbf{1.831}  & \textbf{0.160}  \\ \hline
\end{tabular}
}
\caption{Experiment on event-based human pose dataset. RPC means the rasterized Point Cloud representation \cite{zhao2021point}.}
\label{table: experiment human}
\end{table}
\subsection{Human Pose Estimation Results}
Human pose estimation is a fundamental task that is widely used in virtual reality and security. It aims to accurately regress the human skeletons' key points. Event cameras are distinguished by their advantages in privacy, high speed, and ability to capture motion without blur in such tasks. 
% We follow the DHP19 official split setting, using S1-S12 for training and S13-S17 for testing, and only two front cameras are used following the raw DHP19 dataset.
Point Cloud representation was demonstrated to be efficient for edge devices, but did not perform well in the MPJPE metric. Furthermore, this work proposed the Rasterized Point Cloud representation (RPC), which retains the original Point Cloud network while additionally introducing the representation transformation time, as shown in \cref{table: experiment preprocess}. SECNet leverages the raw unprocessed Event Cloud to reach the SOTA results of point-based and voxel-based methods. Meanwhile, SECNet has the smallest model size, with GMACs of only 0.160, which is 2\% of Point Transformer, as shown in \cref{table: experiment human}. In addition, it takes only 4.37 ms for the model to infer a sample on the server. Furthermore, the citations corresponding to \cref{table: experiment thu,table: ucf hardware,table: experiment human} are provided in the \cref{appendix: comparison method} and visualization results shown in \cref{section: st feature}.
\begin{table}
\centering
\renewcommand\arraystretch{1}
\scalebox{0.8}{
\begin{tabular}{ccccc}
\hline
Dataset          & \multicolumn{2}{c}{\small DHP19}  & \small N-CARS  & \small UCF101-DVS  \\
Metric            & \small  MPJPE$_{2D}$       & \small  MPJPE$_{3D}$       & Acc.   & Acc.   \\ \hline
w all             & \textbf{6.11}        & \textbf{69.89}       & \textbf{0.947}  & \textbf{0.916 } \\
w/o $\alpha$ &6.17 &70.03 &0.942 &0.914\\
w/o EF-KNN &6.20&70.47 &0.937 &0.915\\
% w/o SFT \& w CNN   &             &            &          &       \\
w/o SFA             & 6.68           & 75.43           & 0.929         &0.892      \\
w/o TFA           & 6.47        & 73.84       & 0.944  & 0.913  \\
Input P & 6.27        & 71.12       & 0.937  & 0.913  \\
w/o P      & 6.19        & 70.30       & 0.944 &  0.911     \\ \hline
\end{tabular}
}
\caption{Ablation study on core modules.}
\label{table: experiment modules}
\vspace{-0.5cm}
\end{table}

% The visualization of results can be found in \cref{section: visualization}.

\subsection{Ablation Study}
\subsubsection{Modules Ablation}
In \cref{table: experiment modules}, ``w/o $\alpha$ means SECNet does not adopt D-FPS. And
``w/o \text{EF-KNN}" means SECNet does not adopt EF-KNN. The performance of the model decreased on all three datasets. ``w/o SFA" and ``w/o TFA" denote that SECNet is not equipped with the Spatial-FA module and Temporal-FA module, respectively. The model showed a significant decrease in the DHP19 dataset, suggesting that temporal features are important for predicting the human skeleton. Input P means SECNet incorporates polarity in the input-level instead of the structural-level.``w/o P" represents the input equal to the Point Cloud representation without polarity. All three datasets have demonstrated that polarity provides additional information.  What's more, the more ablation study can be found in the \cref{section: ablation of sfa,section: runtime}. Overall, all the designed components are indispensable and jointly contribute to the superior prediction capability of our proposed SECNet.
\begin{table}
\centering
\renewcommand\arraystretch{1}
\scalebox{0.8}{
\begin{tabular}{ccccc}
\hline
\small Event number & \small SECNet & \small SECNet w LSTM &\small PointNet++ & \small TTPOINT\\ \hline
512          & 46.29  & 44.16   &37.63 &39.70\\
1024         & 50.36  & 48.31  & 43.54 & 42.15\\
2048         & 52.24  & 49.10  &47.02  &42.58\\
4096         & 56.57  & 50.84  &49.04 & 44.39\\
8192         & 60.04  & 53.63  &53.15 &43.65\\ 
10240        & 60.60 &52.42 &54.39 &45.61 \\\hline
\end{tabular}
}
\caption{Ablation study of different Point-based networks in N-Caltech101 with a 300 ms sliding window.}
\label{table: experiment number}
\vspace{-0.5cm}
\end{table}
\subsubsection{Scalability to Event Number}
PointNet++ and TTPOINT do not have an explicit temporal feature extraction module; the accuracy is much lower than that of SECNet, and the performance of TTPOINT does not improve clearly after 4096 points, as shown in \cref{table: experiment number}. Additionally, SECNet replaces LSTM for temporal feature extraction, which makes it difficult to capture long-term dependencies between events as the number of points grows. 
% This demonstrates that SECNet has strong scalability with respect to the number of input points and possesses effective spatio-temporal feature extraction capabilities.
In contrast, the proposed Temporal-FA module models global temporal relationships in the frequency domain, enabling SECNet to effectively capture long-range dependencies while maintaining stable scalability as the event number increases. Moreover, the consistent performance improvement under larger event inputs indicates that SECNet can better utilize the fine-grained temporal information contained in dense Event Cloud representations.
\subsection{Hardware Efficiency and Edge Deployability}
we perform measured FPGA experiments on a Zynq UltraScale+ MPSoC running at 200 MHz to evaluate edge-oriented deployment characteristics. We compare the proposed SFA module against the conventional spatial MLP implementation used for feature extraction.

As shown in Table~\ref{tab:fpga_compare}, the spatial MLP requires 6,302,723 / 12,595,715 / 25,191,430 cycles for stages 1/2/3, corresponding to 31.51 / 62.98 / 125.96 ms latency, respectively. In contrast, the proposed FFT-based SFA accelerator requires only 393,859 cycles (1.97 ms), and the cost remains nearly constant across different stages. This corresponds to approximately $16\times$, $32\times$, and $64\times$ fewer cycles compared with the spatial baseline. Additionally, the on-chip memory RAM18 is reduced from 84 s to 39. This demonstrate SECNet is not only algorithmic lightweight, but also hardware-efficient realizability.

\begin{table}[t]
\centering
\caption{FPGA comparison between spatial MLP and SFA on Zynq UltraScale+ MPSoC (200 MHz).}
\label{tab:fpga_compare}
\resizebox{1\linewidth}{!}{
\begin{tabular}{lcc}
\toprule
Input Dimension & MLP (Cycle Count / Latency) & SFA (Cycle Count / Latency) \\
\midrule
Stage 1 [1024,24,64] & 6,302,723 / 31.51 ms & 393,859 / 1.97 ms \\
Stage 2 [512,24,128] & 12,595,715 / 62.98 ms & 393,859 / 1.97 ms \\
Stage 3 [256,24,256] & 25,191,430 / 125.96 ms & 393,859 / 1.97 ms \\
\midrule
On-Chip Memory (RAMB18) & 84 & 39 \\
\bottomrule
\end{tabular}
}
\end{table}

% These results demonstrate that SECNet is not only lightweight in theory, but also highly hardware-friendly for practical edge deployment.

\section{Conclusion}
In this paper, we carry forward the Event Cloud representation, which is the nearest one to the raw events. We propose event-based G\&S modules and embrace frequency domain analysis to extract long sequence events' features. We benchmark SECNet on ten datasets, and the ground-breaking results demonstrate scalability, efficiency, and effectiveness.

\section*{Acknowledgements}
This work was supported in part by the Youth Science and Technology Talent Support Program of GDSTA (SKXRC2025460), the Guangdong Science and Technology Program (2025A0505000036), the Guangdong Basic and Applied Basic Research Foundation (2026A1515010184), the National Science Foundation of China [U23A20389, 62306095, 82441009, 82441008, 62506101], the Heilongjiang Natural Science Foundation of China [LH2024F017], the China Postdoctoral Science Foundation [2024M764190], and the Fundamental Research Funds for the Central Universities [HIT.NSFJG202439, HIT.NSFJG202434].

\section*{Impact Statement}
This paper aims to advance lightweight and scalable event-based framework for edge applications. By leveraging Event Cloud representation and efficient frequency-domain feature extraction, SECNet enables low-latency and low-power event processing, which is particularly beneficial for resource-constrained scenarios such as robotics, wearable devices, autonomous systems, and smart sensing platforms.

% Compared with conventional frame-based pipelines, our method reduces preprocessing overhead and computational complexity while preserving fine-grained temporal information. This may contribute to more energy-efficient and real-time AI systems in practical deployments, especially for high-speed and privacy-sensitive perception tasks.

As with other event-based perception technologies, the proposed method could potentially be applied in surveillance or monitoring systems. However, SECNet itself is a general-purpose vision backbone and does not introduce additional privacy risks beyond existing event-camera technologies. We hope this work encourages further research toward efficient, reliable, and responsible edge AI systems.

% \begin{algorithm}[tb]
%   \caption{Bubble Sort}
%   \label{alg:example}
%   \begin{algorithmic}
%     \STATE {\bfseries Input:} data $x_i$, size $m$
%     \REPEAT
%     \STATE Initialize $noChange = true$.
%     \FOR{$i=1$ {\bfseries to} $m-1$}
%     \IF{$x_i > x_{i+1}$}
%     \STATE Swap $x_i$ and $x_{i+1}$
%     \STATE $noChange = false$
%     \ENDIF
%     \ENDFOR
%     \UNTIL{$noChange$ is $true$}
%   \end{algorithmic}
% \end{algorithm}

% In the unusual situation where you want a paper to appear in the
% references without citing it in the main text, use \nocite
\nocite{langley00}

\bibliography{example_paper}
\bibliographystyle{icml2026}

%%%%%%%%%%%%%%%%%%%%%%%%%%%%%%%%%%%%%%%%%%%%%%%%%%%%%%%%%%%%%%%%%%%%%%%%%%%%%%%
%%%%%%%%%%%%%%%%%%%%%%%%%%%%%%%%%%%%%%%%%%%%%%%%%%%%%%%%%%%%%%%%%%%%%%%%%%%%%%%
% APPENDIX
%%%%%%%%%%%%%%%%%%%%%%%%%%%%%%%%%%%%%%%%%%%%%%%%%%%%%%%%%%%%%%%%%%%%%%%%%%%%%%%
%%%%%%%%%%%%%%%%%%%%%%%%%%%%%%%%%%%%%%%%%%%%%%%%%%%%%%%%%%%%%%%%%%%%%%%%%%%%%%%
\newpage
\appendix
\onecolumn
\section{References to comparative method abbreviations}
\label{appendix: comparison method}
In \cref{table: experiment thu}, which presents the experiment on the $\text{THU}^{\text{E-ACT}}$-50 dataset, we include HMAX SNN \cite{xiao2019event}, Motion SNN \cite{liu2021event}, and EV-ACT \cite{gao2023action}.

In \cref{table: ucf hardware}, the hardware benchmark on the UCF101-DVS dataset, lists C3D \cite{fan2016video}, ResNext-50 \cite{hara2018can}, 3D-ResNet \cite{hara2018can}, I3D \cite{carreira2017quo}, EventLSTM \cite{annamalai2022event}, ECSNet \cite{chen2022ecsnet}, RG-CNN \cite{bi2020graph}, EVSTr \cite{xie2024event}, VMST-Net \cite{liu2023voxel}, TTPOINT \cite{ren2023ttpoint}, and EventMamba \cite{ren2025rethinking}.

In \cref{table: experiment human}, the experiment on the event-based human pose dataset. We include Pose-Res50 \cite{xiao2018simple}, LeVit-128S \cite{graham2021levit}, DHP19 \cite{calabrese2019dhp19}, VMV-PointTrans \cite{xie2022vmv,zhao2021point}, VMST-Net \cite{liu2023voxel}, PointNet \cite{chen2022efficient,qi2017pointnet}, DGCNN \cite{chen2022efficient,wang2021object}, and Point Transformer \cite{chen2022efficient,zhao2021point}.
\section{Comparison of Event Cloud and Voxel Representations}
\label{section: Comparison}
While Voxel-based (voxel grid and voxel graph) representations can indeed encode a high density of events and preserve polarity through multi-channel binning, they inherently impose a grid-like quantization over the spatiotemporal domain. This discretization leads to two limitations: (1) loss of temporal precision due to uniform binning, and (2) inefficiencies in data transformation and memory usage, especially for high-resolution or long-duration events. In contrast, Event Cloud maintains a continuous representation of each event with no spatial or temporal quantization, preserving the raw structure and allowing direct modeling without costly pre-processing time. Voxel-based representation often requires longer preprocessing time and introduces additional time for constructing graph structures.

To further compare from the network architecture perspective, voxel-based models such as VMV-GCN \cite{xie2022vmv} and VMST-Net \cite{liu2023voxel} typically rely on static receptive fields or geometry-based neighborhood construction, which are limited in their ability to adapt to dynamic spatial-temporal patterns. Their grouping mechanisms are predefined by voxel grid boundaries or spatial distances, ignoring the learned feature similarity between tokens. Moreover,\textbf{ polarity information, although present at the input level, is only utilized in voxel-wise feature encoding, without participating in structural decisions such as neighborhood formation or coordinate update}.

In contrast, SECNet introduces a feature-based G\&S mechanism that leverages both spatial-temporal positions and polarity in a unified 4D feature space. The D-FPS and EF-KNN enable the network to construct adaptive neighborhoods, while CES allows the model to dynamically adjust token positions layer-by-layer based on learned representations. This architectural design not only enhances the semantic alignment of local regions but also improves scalability to large and dense event streams by avoiding fixed voxel grids. Furthermore, SECNet uniquely integrates polarity information into its structural pipeline, making it the only model among the compared methods that fully exploits polarity for both feature and topology modeling. Specific comparison can be seen in  \cref{tab:polarity_comparison_transposed}.
\begin{table}
\centering
\caption{Different Model Architecture Comparison}
\label{tab:polarity_comparison_transposed}
\scalebox{0.65}{
\begin{tabular}{cccccc}
\hline
\textbf{Aspect} & \textbf{SECNet} & PointNet++ \cite{qi2017pointnet++}  &EventNet\cite{sekikawa2019eventnet}& VMV-GCN\cite{xie2022vmv} & VMST-Net\cite{liu2023voxel}\\
\hline
% Polarity Usage Scope & Full: $(x,y,t,p)$ with learnable weights & Patch encoding within voxel only & Channel-separated voxel frame \\
Scalability  & 10240 & 1024  &-&2048 &1024\\
Representation & Event Cloud & Point Cloud  &Event Cloud& Voxel Graph + Muti-view & Voxel Grid \\
Preprocessing Time & short & short  &short& long & long\\
Polarity & Structural-level Integration & \xmark  &Input-level Inclusion& Input-level Inclusion & Input-level Inclusion\\
Sampling Strategy & D-FPS  & FPS  &\xmark& Motion-sensitive & Fixed-grid \\
KNN Distance & Feature-based & Radius-based  &\xmark& Geometry-based & Radius-based \\
Coord. Update & \checkmark (CES)& \xmark  &\xmark& \xmark & \xmark \\
Long Sequence Model& \checkmark & \xmark  &\xmark&\xmark & \xmark \\
\hline
\end{tabular}}
\end{table}

\section{Implement Details}
\label{section: implement details}
For all datasets, the hierarchy structure is configured with 3 loops, and the embedding layer is implemented using a single layer of one-dimensional convolution, with input and output dimensions of 4 and 64, respectively. So the dimensional transformation of Event Cloud features is gradually increased to 64, 132, 268, and 540 as the loop goes on, and the formula is calculated as the previous layer's dimension multiplied by two plus four. As for the number of groups, which is also the number of centroids in the first loop, NMNIST, N-CARS, DVS128 Gesture, and DHP19 are set to 512, ASL-DVS, Daily DVS, and UCF101-DVS are set to 1024, and CIFAR10-DVS and N-Caltech101 are set to 2048. The number of groups is halved as each loop moves to the next one.

\section{Dataset}
\label{section: dataset}
We validate SECNet on three tasks ten datasets, e.g., \textbf{Object Classification}: NMNIST  \cite{orchard2015converting}, N-Caltech101 \cite{orchard2015converting}, CIFAR10-DVS \cite{li2017cifar10}, N-CARS   \cite{sironi2018hats}, ASL-DVS  \cite{bi2020graph}, \textbf{Action Recognition}: DVS128 Gesture \cite{amir2017low}, Daily DVS \cite{liu2021event}, UCF101-DVS \cite{bi2020graph}, $\text{THU}^{\text{E-ACT}}\text{-50}$ \cite{gao2023action}, \textbf{Human Pose Estimation}: DHP19 \cite{calabrese2019dhp19}. For N-Caltech101, CIFAR10-DVS, ASL-DVS, Daily DVS, and UCF101-DVS do not have official divisions, we follow the setting in \cite{bi2020graph,chen2022ecsnet}, randomly select 20\% of the samples as the testset, and all other datasets use official divisions. The details of these datasets are shown in \cref{table: dataset details}.

\section{Super-linear Increase} 
\label{section: super}
PointNet exhibits a linear increase in MACs with the number of points, as it only extracts global information, which is insufficient for complex tasks. In fact, mainstream models such as PointNet++, PointMLP, PointTransformer, and our SECNet extract local features and are also influenced by the number of FPS points (i.e., the number of groups), leading to superlinear growth in computational complexity as shown in \cref{table: lookup table}. Still, SECNet decreases the MACs by 5x given the same sample points.
\begin{table*}
\centering
\renewcommand\arraystretch{1.2}
\scalebox{0.75}{
\begin{tabular}{c|cccccccccccc}
\hline
\multirow{2}{*}{Events Number\textbackslash Centroid} & \multicolumn{3}{c}{[2048, 1024, 512]} & \multicolumn{3}{c}{[1024, 512, 256]} & \multicolumn{3}{c}{[512, 256, 128]} & \multicolumn{3}{c}{[256,128,64]} \\
                  & Runtime     & Param.     & GMACs    & Runtime     & Param.    & GMACs    & Runtime    & Param.    & GMACs    & Runtime    & Param.    & GMACs   \\ \hline
10240             & 19.71       & 1.514      & 1.198    & 9.70        & 1.103     & 0.548    & 5.08       & 0.898     & 0.262    & 4.43       & 0.795     & 0.129   \\
8192              & 16.8        & 1.514      & 1.063    & 8.34        & 1.103     & 0.48     & 4.37       & 0.898     & 0.228    & 4.43       & 0.795     & 0.112   \\
4096              & 10.8        & 1.514      & 0.794    & 5.30        & 1.103     & 0.345    & 4.43       & 0.898     & 0.160    & 4.40       & 0.795     & 0.077   \\
2048              & -           & -          & -        & 4.44        & 1.103     & 0.278    & 4.38       & 0.898     & 0.126    & 4.35       & 0.795     & 0.060   \\
1024              & -           & -          & -        & -           & -         & -        & 4.35       & 0.898     & 0.109    & 4.35       & 0.795     & 0.051   \\ \hline
\end{tabular}
}
\caption{Hardware performance lookup table for SECNet.}
\label{table: lookup table}
\end{table*}

\section{Specific Dimension Variance in SECNet}
\label{section: specific dimension}
\begin{figure*}
  \centering
  \includegraphics[width=0.6\linewidth]{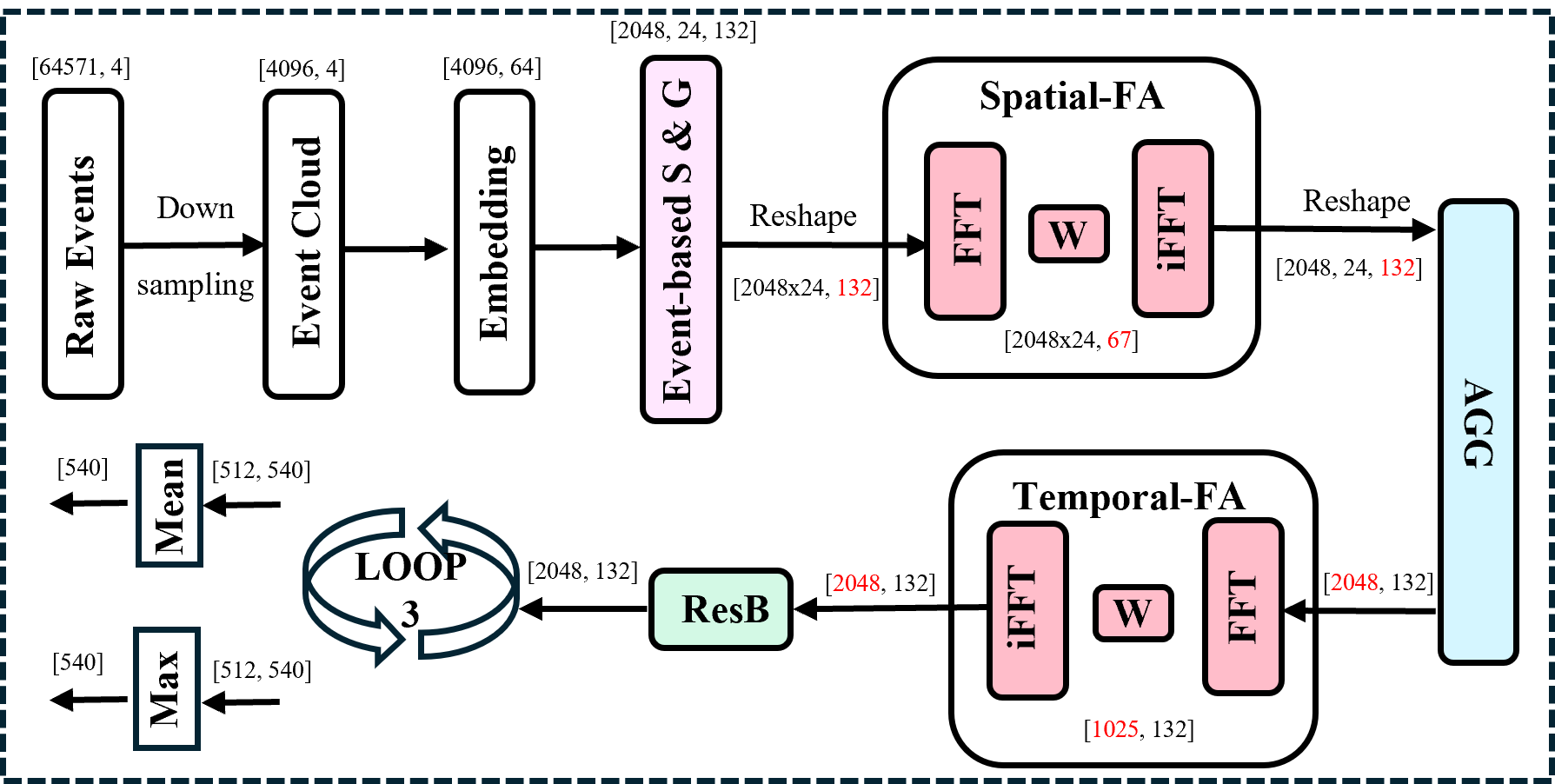}
   \caption{Specific variation of dimensions of a sample in SECNet.}
   \label{fig:onecol}
\end{figure*}
We show the specific variation of the dimension of a sample in \cref{fig:onecol}. The red marks represent the dimensions done FFT.
1. Spatial-FA: This module functions on channel dimension (132), which represents spatial information between each event point and the central feature. Unlike the frame representation, the feature does not have x and y dimensions but rather abstracts the information of x and y into the channel dimension. 2. Temporal-FA: This module functions on the event number dimension (2048), which represents discretized temporal information due to its being strictly arranged in the order of events emitted. 

\section{Hardware Resource Performance}
\label{section: hardware performance}
In addition, we tested the hardware resource performance and runtime of SECNet on the server platform. 
% \cref{table: gesture hardware} shows the amazing results of SECNet. 
Thanks to the design of the Spatial-FA module, SECNet achieves leading results with a remarkably low computational cost, requiring only one fortieth of the MACs compared to frame-based methods. Additionally, SECNet inherits the high inference speed of Point Cloud representations, delivering an impressive throughput of 229.7 FPS while far superior to frame-based and voxel-based methods. Although the number of parameters in SECNet is higher than that of TTPOINT, the tensor decomposition technique employed significantly reduces the network's inference time. SECNet still maintains a relatively low parameter count compared to the frame-based method.

\section{Pseudo Code}
\label{section: psedo}
The SECNet pipeline is shown in \cref{algorithm}. The input is an Event Cloud along with the stage number parameter $m$. The output consists of the predicted category $\hat{c}$ and human pose $\hat{p}$.

First, the algorithm embeds the input Event Cloud using the embedding module, producing the feature representation $\mathcal{F}$. Then, the algorithm proceeds through a series of stages ($m$ iterations), where each stage consists of several modules. The Event-based G \& S module processes the input $\mathcal{EC}$ and the embedded features $\mathcal{F}$ via the G \& S function, generating group set $\mathcal{G}$, group features $\mathcal{F}_{\mathcal{G}}$, and a modified Event Cloud $\mathcal{EC}'$. Next, the Spatial Frequency-aware (SFA) module aggregates the feature graph $\mathcal{F}_{\mathcal{G}}$ spatially to produce the spatially aggregated features $\mathcal{F}_{\mathcal{S}}$. Afterward, the Temporal Aggregate module temporally aggregates the spatially aggregated features $\mathcal{F}_{\mathcal{S}}$ to produce the temporally aggregated features $\mathcal{F}_{A}$. Then, the Temporal Feature Aggregation (TFA) module processes the temporally aggregated features $\mathcal{F}_{A}$ to generate the temporal features $\mathcal{F}_{\mathcal{T}}$. Finally, the ResB module applies residual learning to the temporal features $\mathcal{F}_{\mathcal{T}}$ to obtain the final features $\mathcal{F}_{\mathcal{R}}$.

After each stage, the feature set $\mathcal{F}$ is updated with the residual features $\mathcal{F}_{\mathcal{R}}$, and the Event Cloud $\mathcal{EC}$ is updated with the modified Event Cloud $\mathcal{EC}'$. Upon completion of all stages, the algorithm computes the mean and maximum of the feature set $\mathcal{F}$ and combines them using a concat operation $\oplus$. Finally, a classifier regressor is applied to the fused features to predict the category $\hat{c}$ and extract 13 skeleton points $\hat{p}$ corresponding to the human pose.

This pipeline involves a series of feature transformations and aggregations aimed at efficiently and effectively processing event-driven data for both category classification and human pose estimation.
\begin{algorithm}[t]
\caption{SECNet Pipeline}
\textbf{Input}: Event Cloud $\mathcal{EC}$\\
\textbf{Parameters}: Stage Number: $m$\\
\textbf{Output}: Category $\hat{c}$, Human Pose $\hat{p}$
\begin{algorithmic}[1] %[1] enables line numbers
\STATE $\mathcal{F} = \text{Embedding}(\mathcal{EC}),$
\FOR{ i in $\text{stage}(m)$}
% \STATE \textbf{Grouping and Sampling}($PN$)
% \STATE Get $PGS \in [B, T'_{i}, K, 2*D_{\text{stage-1}}]$
\STATE \textbf{Event-based G \& S Module} 
\STATE $\mathcal{G},\mathcal{F}_{\mathcal{G}},\mathcal{EC}' = \text{G\&S}(\mathcal{EC},\mathcal{F})$
\STATE \textbf{Spatial-FA Module}
\STATE $\mathcal{F}_{\mathcal{S}} = \text{SFA}(\mathcal{F}_{\mathcal{G}})$;
\STATE \textbf{Temporal Aggregate Module}
\STATE $\mathcal{F}_{A} = \text{AGG}(\mathcal{F}_{\mathcal{S}})$;
\STATE \textbf{Temporal-FA Module}
\STATE $\mathcal{F}_{\mathcal{T}} = \text{TFA}(\mathcal{F}_{A})$;
\STATE \textbf{ResB Module}
\STATE $\mathcal{F}_{R} = \text{RES}(\mathcal{F}_{T})$
\STATE $\mathcal{F} = \mathcal{F}_{R}$; $\mathcal{EC} = \mathcal{EC}'$;
\ENDFOR
\STATE $\mathcal{F} = \text{Mean}(\mathcal{F})\oplus\text{Max}{(\mathcal{F})}$
\STATE \textbf{Classifier} \quad \quad \quad \textbf{Regressor}
\STATE Get category $\hat{c}$ \quad Get 13 skeleton points $\hat{p}$
% \STATE \textbf{Regressor} 
% \STATE Get 13 skeleton points $\hat{p}$
\end{algorithmic}
\label{algorithm}
\end{algorithm}
\section{Ablation Study on Spatial-FA Module}
\label{section: ablation of sfa}
\begin{figure}
\centerline{\includegraphics[width= 8cm]{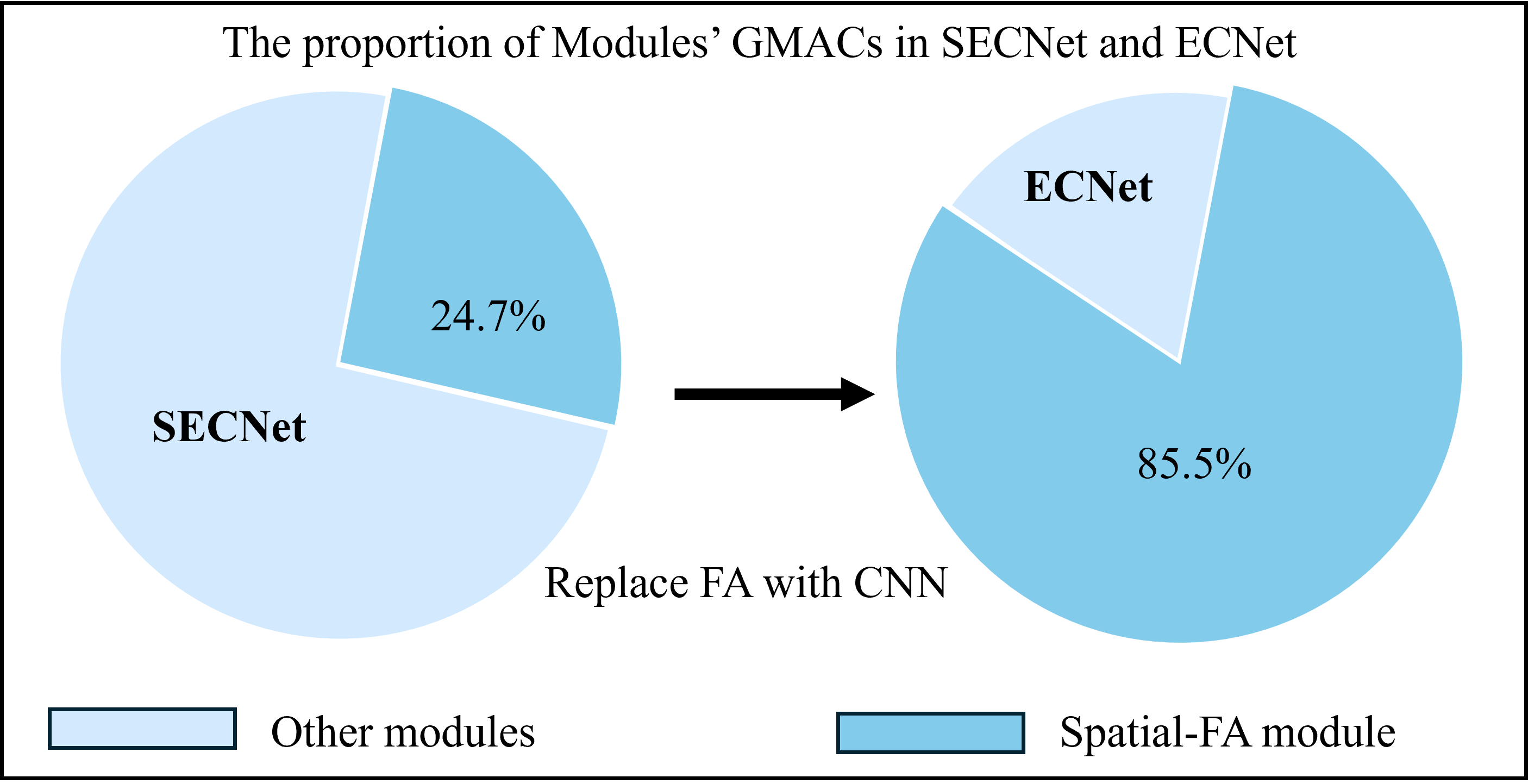}}
\caption{Proportional chart of GMACs. The other modules' GMACs, represented by the light blue, are congruent.}
\label{fig: proportional chart}
\end{figure}

As mentioned in the main paper, we replace the convolution operations with Spatial-FA, and this significantly reduces the overall GMACs of the model.
To explore its effect on the results, we utilize convolution to extract spatial features on six datasets and list the metric change in \cref{table: sfa ablation} \cite{calabrese2019dhp19,bi2020graph,sironi2018hats,orchard2015converting,amir2017low}. ECNet denotes the network in which the Spatial-FA module of SECNet is replaced with a traditional convolution. The proportion of overall GMACs accounted for by the Spatial-FA module before and after the replacement is shown in \cref{fig: proportional chart}. 
Although SECNet generally underperforms ECNet across most datasets, it is markedly more efficient, with its GMACs constituting only 5\% of the size of ECNet and its model parameters being smaller. Furthermore, SECNet has surpassed nearly all other representations in terms of performance.
\section{Look up Table for Runtime}
\label{section: runtime}
Differences in the number of input events and the centroid of each loop in SECNet can influence its hardware performance. We evaluated the hardware performance of various combinations on the server platform, as shown in \cref{table: lookup table}, measuring runtime in milliseconds (ms), parameters in millions (M), and MACs in giga (G).
\begin{figure*}[t]
\centerline{\includegraphics[width= 13cm]{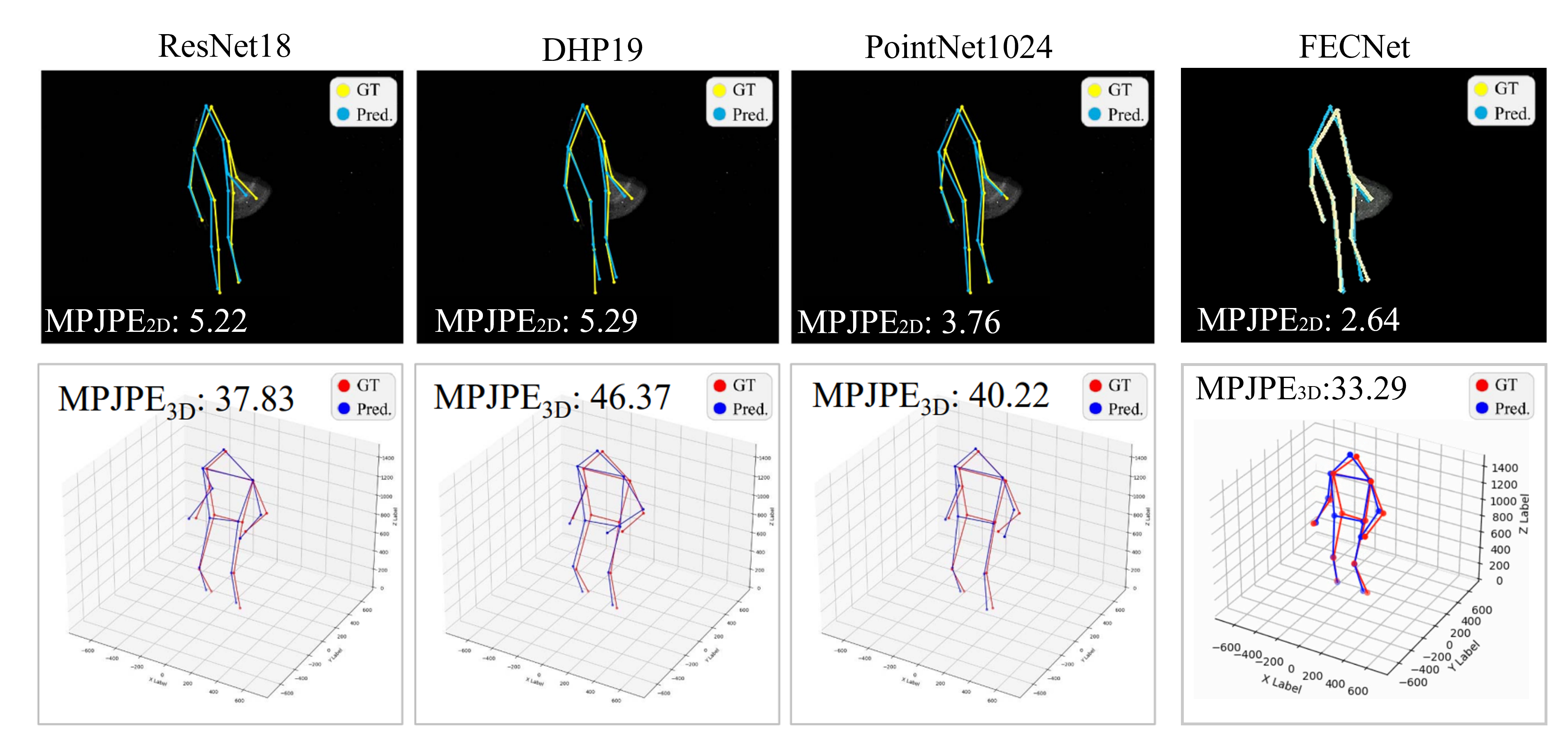}}
\caption{ Human pose estimation results in visualization for different models. The first row shows the results in 2D (yellow for ground truth, blue for prediction). The second row shows the results in 3D (red for ground truth, blue for prediction).}
\label{fig: si compare human pose}
\end{figure*}
By observing the table, we can find that 1. the number of input events is independent of the model size, and the number of centroids per loop solely determines it; 2. The runtime increases synchronously with the event number and centroid. 3. GMAC super-linear grows with the increase of events number and centroid. Additionally, simply increasing the number of events will slow down the network inference, creating a trade-off between performance and inference speed in practical applications. The throughput of SECNet is guaranteed to be above 200 FPS when the centroid is kept at [512, 256, 128] and [256, 128, 64].
\begin{table*}
\centering
\renewcommand\arraystretch{1.2}
\scalebox{0.8}{
\begin{tabular}{ccccccccccccccc}
\hline
Dataset   & \multicolumn{4}{c}{DHP19}                              & \multicolumn{2}{c}{ASLDVS} & \multicolumn{2}{c}{UCF101-DVS} & \multicolumn{2}{c}{NCARS } & \multicolumn{2}{c}{NMNIST} & \multicolumn{2}{c}{DVS128 Gesture} \\
Method    & \multicolumn{2}{c}{SECNet} & \multicolumn{2}{c}{ECNet} & SECNet       & ECNet       & SECNet         & ECNet         & SECNet       & ECNet      & SECNet       & ECNet       & SECNet           & ECNet           \\ \hline
Metric & 6.11        & 69.89        & 5.87        & 66.87       & 0.9993       & 0.9994      & 0.916          & 0.941         & 0.947        & 0.951      & 0.997        & 0.997       & 0.989            & 0.983           \\
Parameter & \multicolumn{2}{c}{1.831}  & \multicolumn{2}{c}{2.193} & 1.107        & 1.469       & 1.126          & 1.489         & 0.896        & 1.258      & 1.103        & 1.465       & 0.898            & 1.26            \\
GMACs     & \multicolumn{2}{c}{0.160}  & \multicolumn{2}{c}{0.904} & 0.345        & 1.832       & 0.48           & 1.967         & 0.228        & 0.971      & 0.345        & 1.832       & 0.11             & 0.903           \\
SFA MMACs & \multicolumn{2}{c}{39.5}   & \multicolumn{2}{c}{783}   & 79           & 1566        & 79             & 1566          & 39.5         & 802        & 79           & 1566        & 39.5             & 783             \\ \hline
\end{tabular}
}
\caption{Ablation study for the Spatial-FA module of SECNet.}
\label{table: sfa ablation}
\end{table*}
\begin{figure}
\centerline{\includegraphics[width= 16cm]{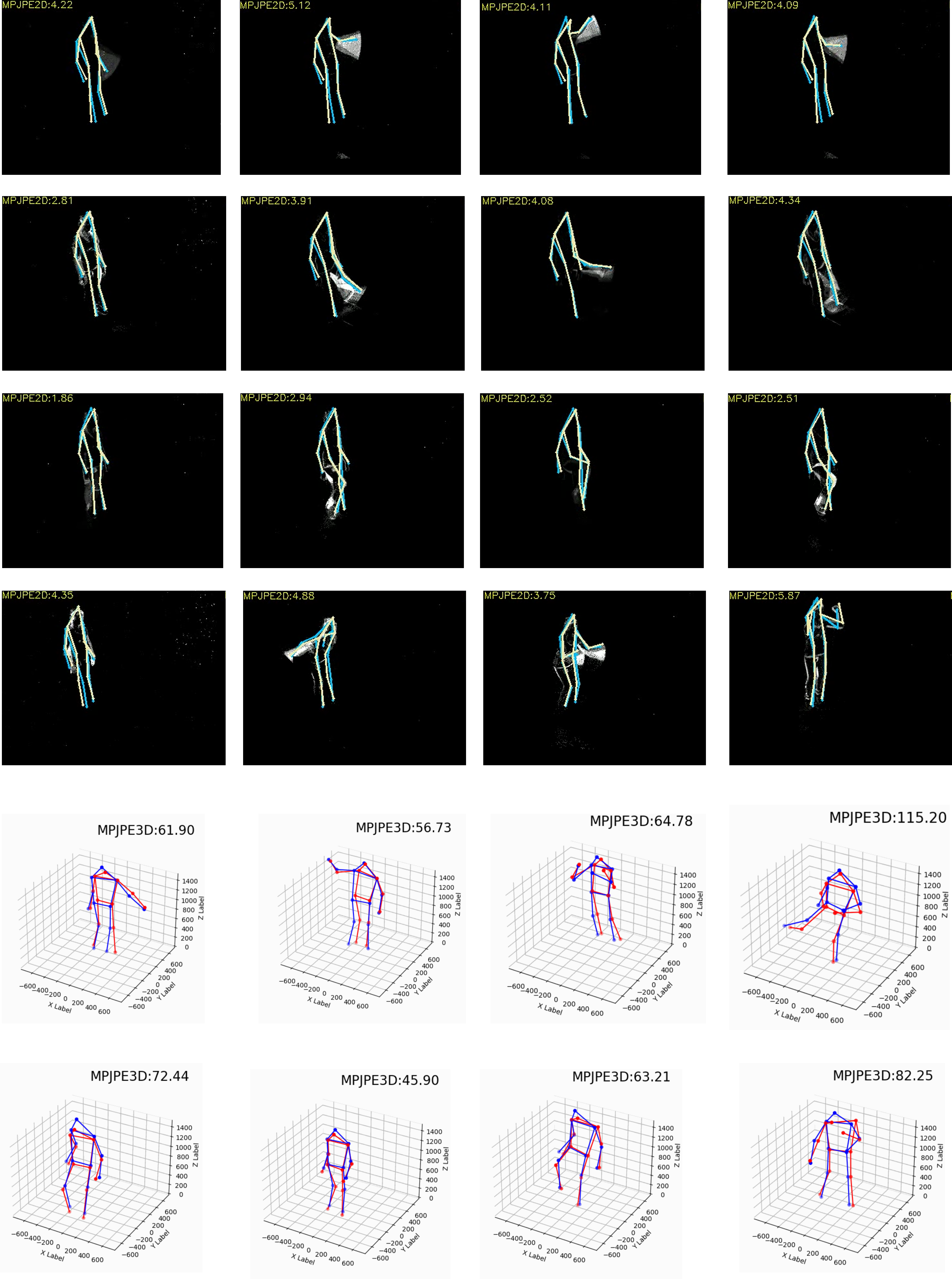}}
\caption{ More 2D and 3D SECNet's results in DHP19 dataset.}
\label{fig: more dhp19 results}
\end{figure}
\section{Spatio-temporal Features}
\label{section: st feature}
Event Cloud contains both spatial $(x, y)$ and temporal $t$ coordinates, and the features extracted through the network using these two types of coordinates refer to spatial-temporal features. Additionally, the manner in which these coordinates are employed results in distinct categories of features.  We break down the categories of temporal features into \textbf{explicit} and \textbf{implicit} types. Previous works have treated the events' $t$ coordinate as the space $z$ coordinate, employing the Point Cloud network to abstract features where $t$ is regarded as an implicit variable analogous to $(x, y)$. So, we call such features implicit temporal features.  In contrast, when $t$ is treated as an index within networks such as LSTM \cite{sherstinsky2020fundamentals}, SSM \cite{gu2023mamba}, and attention mechanisms \cite{niu2021review}, we classify the resulting features as explicit temporal features.

\section{Visualization}
\label{section: visualization}
% We visualize the serval DHP19 results generated by SECNet. And 
\cref{fig: si compare human pose} presents the prediction results of various models applied to the same test samples, whose settings are consistent with the paper \cite{chen2022efficient}. ResNet18 and DHP19 employ the event frame representation \cite{calabrese2019dhp19}, PointNet utilizes the rasterized Point Cloud representation \cite{chen2022efficient}, and SECNet adopts the Event Cloud representation. The results of SECNet are superior to those of the other three methods. More results for different sequences in the dataset are shown in \cref{fig: more dhp19 results}.
% \section{You \emph{can} have an appendix here.}

% You can have as much text here as you want. The main body must be at most $8$
% pages long. For the final version, one more page can be added. If you want, you
% can use an appendix like this one.

% The $\mathtt{\backslash onecolumn}$ command above can be kept in place if you
% prefer a one-column appendix, or can be removed if you prefer a two-column
% appendix.  Apart from this possible change, the style (font size, spacing,
% margins, page numbering, etc.) should be kept the same as the main body.
%%%%%%%%%%%%%%%%%%%%%%%%%%%%%%%%%%%%%%%%%%%%%%%%%%%%%%%%%%%%%%%%%%%%%%%%%%%%%%%
%%%%%%%%%%%%%%%%%%%%%%%%%%%%%%%%%%%%%%%%%%%%%%%%%%%%%%%%%%%%%%%%%%%%%%%%%%%%%%%

\end{document}